\documentclass[10pt,twocolumn,letterpaper]{article}

\usepackage{cvpr}
\usepackage{times}
\usepackage{epsfig}
\usepackage{graphicx}
\usepackage{amsmath}
\usepackage{amssymb}
\usepackage{enumitem}
\usepackage{array}
\usepackage[breaklinks=true,bookmarks=false,colorlinks,hypertexnames=false]{hyperref}
\usepackage{cleveref}
\usepackage{authblk}
\usepackage{chngcntr}
\usepackage{multirow}

\cvprfinalcopy

\def\cvprPaperID{4790}

\ifcvprfinal\fi
\begin{document}

\title{Learning Texture Transformer Network for Image Super-Resolution}
\author[1$\ast$]{\vspace{-0.5cm}Fuzhi Yang}
\author[2]{Huan Yang}
\author[2]{Jianlong Fu}
\author[1]{Hongtao Lu}
\author[2]{Baining Guo}
\affil[1]{Department of Computer Science and Engineering, \authorcr MoE Key Lab of Artificial Intelligence, AI Institute, Shanghai Jiao Tong University,}
\affil[2]{Microsoft Research, Beijing, P.R. China, \authorcr \{yfzcopy0702, htlu\}@sjtu.edu.cn, \{huayan, jianf, bainguo\}@microsoft.com}

\maketitle
\thispagestyle{empty}

\renewcommand{\thefootnote}{\fnsymbol{footnote}}
\footnotetext[1]{This work was performed when the first author was visiting Microsoft Research as a research intern.}

\begin{abstract}
We study on image super-resolution (SR), which aims to recover realistic textures from a low-resolution (LR) image. Recent progress has been made by taking high-resolution images as references (Ref), so that relevant textures can be transferred to LR images. However, existing SR approaches neglect to use attention mechanisms to transfer high-resolution (HR) textures from Ref images, which limits these approaches in challenging cases. In this paper, we propose a novel \textbf{T}exture \textbf{T}ransformer Network for Image \textbf{S}uper-\textbf{R}esolution (TTSR), in which the LR and Ref images are formulated as queries and keys in a transformer, respectively. TTSR consists of four closely-related modules optimized for image generation tasks, including a learnable texture extractor by DNN, a relevance embedding module, a hard-attention module for texture transfer, and a soft-attention module for texture synthesis. Such a design encourages joint feature learning across LR and Ref images, in which deep feature correspondences can be discovered by attention, and thus accurate texture features can be transferred. The proposed texture transformer can be further stacked in a cross-scale way, which enables texture recovery from different levels (e.g., from $1\times$ to $4\times$ magnification). Extensive experiments show that TTSR achieves significant improvements over state-of-the-art approaches on both quantitative and qualitative evaluations. The source code can be downloaded at \url{https://github.com/researchmm/TTSR}.
\end{abstract}

\begin{figure}[!t]
    \centering
    \includegraphics[width=\linewidth, page=1]{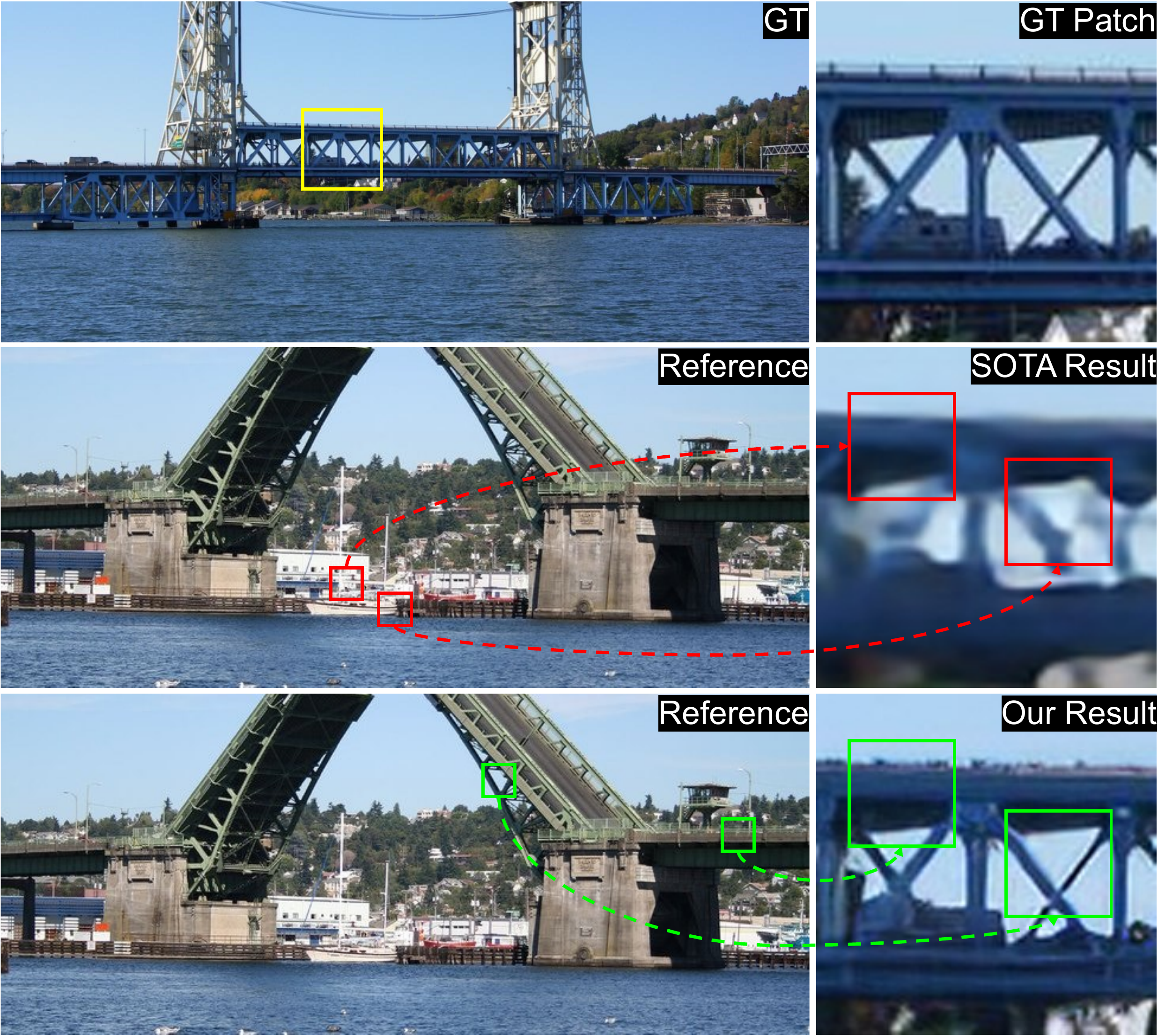}
    \caption{A comparison of $4\times$ SR results between the proposed TTSR and a state-of-the-art RefSR approach~\cite{zheng2018crossnet}. TTSR (ours) learns to search on relevant textures from the Ref image (indicated by green) for a target LR region (indicated by yellow), which avoids incorrect texture transfer (indicated by red).}
    \label{fig:teaser}
\end{figure}

\section{Introduction}
Image super-resolution aims to recover natural and realistic textures for a high-resolution image from its degraded low-resolution counterpart~\cite{IraniImproving}. The recent success of image SR can greatly enhance the quality of media content for a better user experiences. For example, the digital zoom algorithm for mobile cameras and image enhancement technology for digital televisions. Besides, this fundamental technology can benefit a broad range of computer vision tasks, like medical imaging~\cite{oktay2016multi} and satellite imaging~\cite{yildirim2012novel}.

The research on image SR is usually conducted on two paradigms, including single image super-resolution (SISR), and reference-based image super-resolution (RefSR). Traditional SISR often results in blurry effects, because the high-resolution (HR) textures have been excessively destructed in the degrading process which are unrecoverable. Although generative adversarial networks (GANs)~\cite{goodfellow2014generative} based image SR approaches are proposed to relieve the above problems, the resultant hallucinations and artifacts caused by GANs further pose grand challenges to image SR tasks.

Recent progress has been made by reference-based image super-resolution (RefSR), which transfers HR textures from a given Ref image to produce visually pleasing results~\cite{freedman2011image, freeman2002example, sun2012super, timofte2013anchored, yue2013landmark}. However, state-of-the-art (SOTA) approaches usually adopt a straightforward way to transfer textures which may result in unsatisfied SR images (as shown in Figure.~\ref{fig:teaser}). For example, Zheng et al.~\cite{zheng2018crossnet} adopts a flow-based approach which usually searches and transfers inaccurate textures (indicate by red) when facing large viewpoint changes between the LR and Ref image. Zhang et al.~\cite{zhang2019image} adopts a feature space defined by a pre-trained classification model to search and transfer textures between the LR and Ref image. Nevertheless, such high-level semantic features can not effectively represent HR textures which remain to generate implausible results.

To address these problems, we propose a novel \textbf{T}exture \textbf{T}ransformer Network for Image \textbf{S}uper-\textbf{R}esolution (TTSR). Specifically, four closely-related modules optimized for image generation tasks are proposed. First, we propose a learnable texture extractor, in which parameters will be updated during end-to-end training. Such a design enables a joint feature embedding of LR and Ref images which creates a solid foundation for applying attention mechanism~\cite{ma2018gan, xu2018attngan, vaswani2017attention} in SR tasks. Second, we propose a relevance embedding module to compute the relevance between the LR and Ref image. More specifically, we formulate the extracted features from the LR and Ref image as the query and key in a transformer~\cite{vaswani2017attention} to obtain a hard-attention map and a soft-attention map. Finally, we propose a hard-attention module and a soft-attention module to transfer and fuse HR features from the Ref image into LR features extracted from backbone through the attention maps. The design of TTSR encourages a more accurate way to search and transfer relevant textures from Ref to LR images.

Furthermore, we propose a cross-scale feature integration module to stack the texture transformer, in which the features are learnt across different scales (e.g, from $1\times$ to $4\times$) to achieve a more powerful feature representation. As shown in Figure~\ref{fig:teaser}, the overall design enables our TTSR to search and transfer relevant textures from the Ref image (indicated by green) which achieves a better visual result compared with SOTA approaches. The main contributions of this paper are:
\begin{itemize}[nosep]
  \item To the best of our knowledge, we are one of the first to introduce the transformer architecture into image generation tasks. More specifically, we propose a texture transformer with four closely-related modules for image SR which achieves significant improvements over SOTA approaches.
  \item We propose a novel cross-scale feature integration module for image generation tasks which enables our approach to learn a more powerful feature representation by stacking multiple texture transformers.
\end{itemize}

\section{Related Work}\label{sec:related}
In this section, we review previous works of single image super-resolution (SISR) and reference-based image super-resolution (RefSR) which are the most relevant to our work.

\subsection{Single Image Super-Resolution}
In recent years, deep learning based SISR methods have achieved significant improvements over traditional non-learning based methods. Deep learning based methods in SISR treat this problem as a dense image regression task which learns an end-to-end image mapping function represented by a CNN between LR and HR images. 

Dong et al.~\cite{dong2015image} proposed SRCNN that firstly adopted deep learning into SISR by using a three-layer CNN to represent the mapping function. Dong et al.~\cite{dong2016accelerating} further sped up the SR process by replacing the interpolated LR image with the original LR image and using deconvolution at the very last layer to enlarge the feature map. Soon afterwards, Kim et al. proposed VDSR~\cite{kim2016accurate} and DRCN~\cite{kim2016deeply} with deeper networks on residual learning. Shi et al.~\cite{shi2016real} replaced deconvolution with the subpixel convolution layer to reduce the checkerboard artifact. Residual block~\cite{he2016deep} was introduced into SISR in SRResNet~\cite{ledig2017photo} and improved in EDSR~\cite{lim2017enhanced}. With the help of residual block, a lot of works focused on designing deeper or wider networks~\cite{dai2019second, tai2017image, tai2017memnet}. Zhang et al.~\cite{zhang2018residual} and Tong et al.~\cite{tong2017image} adopted dense blocks~\cite{huang2017densely} to combine features from different levels. Zhang et al.~\cite{zhang2018image} improved residual block by adding channel attention. Liu et al.~\cite{liu2018non} proposed a non-local recurrent network for image restoration. Dai et al.~\cite{dai2019second} introduced second-order statistics for more discriminative feature representations. 

The above methods use mean square error (MSE) or mean absolute error (MAE) as their objective function which ignores human perceptions. In recent years, more and more works aim to improve perceptual quality. Johnson et al.~\cite{johnson2016perceptual} introduced perceptual loss into SR tasks, while SRGAN~\cite{ledig2017photo} adopted generative adversarial networks (GANs)~\cite{goodfellow2014generative} and showed visually satisfying results. Sajjadi et al.~\cite{sajjadi2017enhancenet} used Gram matrix based texture matching loss to enforce local similar textures, while ESRGAN~\cite{wang2018esrgan} enhanced SRGAN by introducing RRDB with relativistic adversarial loss. Recent proposed RSRGAN~\cite{zhang2019ranksrgan} trained a ranker and used rank-content loss to optimize the perceptual quality, which achieved state-of-the-art visual results.

\subsection{Reference-based Image Super-Resolution}
Different from SISR, RefSR can harvest more accurate details from the Ref image. This could be done by several approaches like image aligning or patch matching. Some existing RefSR approaches~\cite{wang2016light, yue2013landmark, zheng2018crossnet} choose to align the LR and Ref image. Landmark~\cite{yue2013landmark} aligned the Ref image to the LR image through a global registration to solve an energy minimization problem. Wang et al.~\cite{wang2016light} enhanced the Ref image by recurrently applying non-uniform warping before feature synthesis. CrossNet~\cite{zheng2018crossnet} adopted optical flow to align the LR and Ref image at different scales and concatenated them into the corresponding layers of the decoder. However, the performance of these methods depends largely on the aligning quality between the LR and Ref image. Besides, the aligning approaches such as optical flow are time-consuming, which is adverse to real applications. 

Other RefSR approaches ~\cite{boominathan2014improving, zhang2019image, zheng2017learning} adopt ``patch match'' method to search proper reference information. Boominathan et al.~\cite{boominathan2014improving} matched the patches between gradient features of the LR and down-sampled Ref image. Zheng. et al.~\cite{zheng2017learning} replaced the simple gradient features with features in convolution neural networks to apply semantic matching and used a SISR method for feature synthesis. Recent work SRNTT~\cite{zhang2019image} applied patch matching between VGG~\cite{simonyan2014very} features of the LR and Ref image to swap similar texture features. However, SRNTT ignores the relevance between original and swapped features and feeds all the swapped features equally into the main network.

To address these problems, we propose a texture transformer network which enables our approach to search and transfer relevant textures from Ref to LR images. Moreover, the performance of our approach can be further improved by stacking multiple texture transformers with a proposed cross-scale feature integration module.

\section{Approach}\label{sec:approach}
In this section, we introduce the proposed \textbf{T}exture \textbf{T}ransformer Network for Image \textbf{S}uper-\textbf{R}esolution (TTSR). On top of the texture transformer, we propose a cross-scale feature integration module (CSFI) to further enhance model performances. The texture transformer and CSFI will be discussed in Section \ref{sec:TT} and Section \ref{sec:CSFI}, respectively. A group of loss functions for optimizing the proposed network will be explained in Section \ref{sec:loss}.

\subsection{Texture Transformer}\label{sec:TT}

\begin{figure}[t]
\centering
\includegraphics[width=\linewidth, page=1]{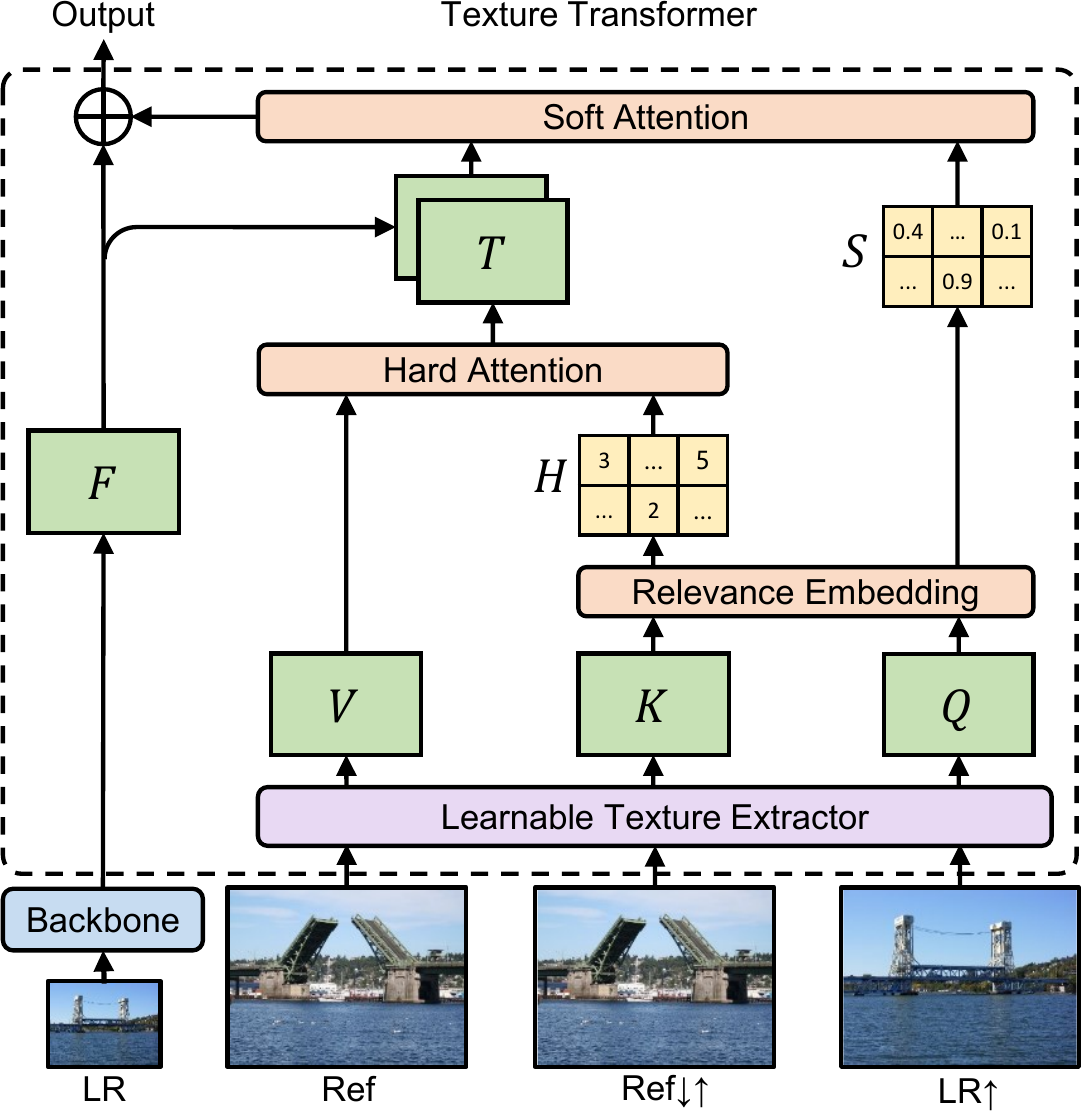}
\caption{The proposed texture transformer. $Q$, $K$ and $V$ are the texture features extracted from an up-sampled LR image, a sequentially down/up-sampled Ref image, and an original Ref image, respectively. $H$ and $S$ indicate the hard/soft attention map, calculated from relevance embedding. $F$ is the LR features extracted from a DNN backbone, and is further fused with the transferred texture features $T$ for generating the SR output.}
\label{fig:texture_transformer}
\end{figure}

The structure of the texture transformer is shown in Figure~\ref{fig:texture_transformer}. LR, LR$\uparrow$ and Ref represent the input image, the $4\times$ bicubic-upsampled input image and the reference image, respectively. We sequentially apply bicubic down-sampling and up-sampling with the same factor $4\times$ on Ref to obtain Ref$\downarrow \uparrow$ which is domain-consistent with LR$\uparrow$. The texture transformer takes Ref, Ref$\downarrow \uparrow$, LR$\uparrow$ and the LR features produced by the backbone as input, and outputs a synthesized feature map, which will be further used to generate the HR prediction. There are four parts in the texture transformer: the learnable texture extractor (LTE), the relevance embedding module (RE), the hard-attention module for feature transfer (HA) and the soft-attention module for feature synthesis (SA). Details will be discussed below.

\noindent\textbf{Learnable Texture Extractor.} In RefSR tasks, texture extraction for reference images is essential because accurate and proper texture information will assist the generation of SR images. Instead of using semantic features extracted by a pre-trained classification model like VGG~\cite{simonyan2014very}, we design a learnable texture extractor whose parameters will be updated during end-to-end training. Such a design encourages a joint feature learning across the LR and Ref image, in which more accurate texture features can be captured. The process of texture extraction can be expressed as:
\begin{align}
    Q &= LTE (LR \uparrow), \\
    K &= LTE (Ref \downarrow \uparrow), \\
    V &= LTE (Ref),
\end{align}
where $LTE(\cdot)$ denotes the output of our learnable texture extractor. The extracted texture features, $Q$ (query), $K$ (key), and $V$ (value) indicate three basic elements of the attention mechanism inside a transformer and will be further used in our relevance embedding module.

\noindent\textbf{Relevance Embedding.}
Relevance embedding aims to embed the relevance between the LR and Ref image by estimating the similarity between Q and K. We unfold both Q and K into patches, denoted as $q_i$ ($i \in [1, H_{LR} \times W_{LR}]$) and $k_j$ ($j \in [1, H_{Ref} \times W_{Ref}]$). Then for each patch $q_i$ in $Q$ and $k_j$ in $K$, we calculate the relevance $r_{i,j}$ between these two patches by normalized inner product:
\begin{align}
    r_{i,j} = \left< \frac{q_{i}}{\left\| q_{i} \right\|}, \frac{k_{j}}{\left\| k_{j} \right\|} \right>.
\end{align}
The relevance is further used to obtain the hard-attention map and the soft-attention map.

\noindent\textbf{Hard-Attention.} We propose a hard-attention module to transfer the HR texture features $V$ from the Ref image. Traditional attention mechanism takes a weighted sum of $V$ for each query $q_i$. However, such an operation may cause blur effect which lacks the ability of transferring HR texture features. Therefore, in our hard-attention module, we only transfer features from the most relevant position in $V$ for each query $q_i$. 

More specifically, we first calculate a hard-attention map $H$ in which the $i$-th element $h_i$ ($i \in [1, H_{LR} \times W_{LR}]$) is calculated from the relevance $r_{i,j}$:
\begin{align}
     h_i &= \mathop{\arg\max}_{j} r_{i,j}.
\end{align}
The value of $h_i$ can be regarded as a hard index, which represents the most relevant position in the Ref image to the $i$-th position in the LR image. To obtain the transferred HR texture features $T$ from the Ref image, we apply an index selection operation to the unfolded patches of $V$ using the hard-attention map as the index:
\begin{align}
    t_i = v_{h_i},
\end{align}
where $t_i$ denotes the value of $T$ in the $i$-th position, which is selected from the $h_i$-th position of $V$.

As a result, we obtain a HR feature representation $T$ for the LR image which will be further used in our soft-attention module.

\noindent\textbf{Soft-Attention.} We propose a soft-attention module to synthesize features from the transferred HR texture features $T$ and the LR features $F$ of the LR image from a DNN backbone. During the synthesis process, relevant texture transfer should be enhanced while the less relevant ones should be relived. To achieve that, a soft-attention map $S$ is computed from $r_{i,j}$ to represent the confidence of the transferred texture features for each position in $T$:
\begin{align}
    s_i &= \mathop{\max}_{j} r_{i,j},
\end{align}
where $s_i$ denotes the $i$-th position of the soft-attention map $S$. Instead of directly applying the attention map $S$ to $T$, we first fuse the HR texture features $T$ with the LR features $F$ to leverage more information from the LR image. Such fused features are further element-wisely multiplied by the soft-attention map $S$ and added back to $F$ to get the final output of the texture transformer. This operation can be represented as:
\begin{align}
    F_{out} = F + Conv(Concat(F, T)) \odot S,
\end{align}
where $F_{out}$ indicates the synthesized output features. $Conv$ and $Concat$ represent a covolutional layer and Concatenation operation, respectively. The operator $\odot$ denotes element-wise multiplication between feature maps.

In summary, the texture transformer can effectively transfer relevant HR texture features from the Ref image into the LR features, which boosts a more accurate process of texture generation.

\begin{figure}[!t]
\centering
\includegraphics[width=\linewidth, page=2]{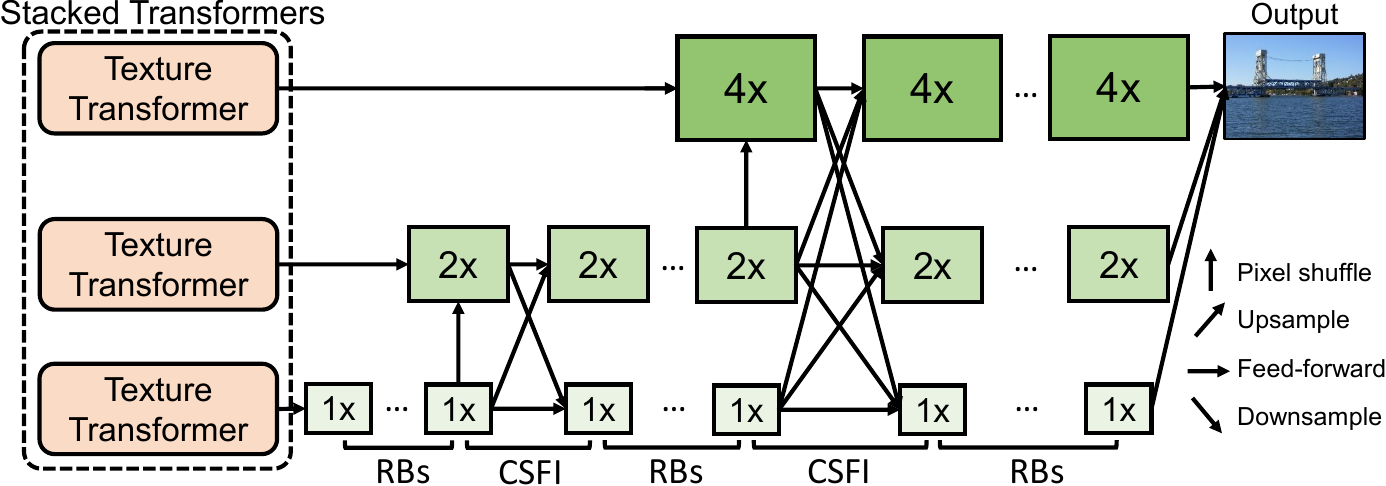}
\caption{Architecture of stacking multiple texture transformers in a cross-scale way with the proposed cross-scale feature integration module (CSFI). RBs indicates a group of residual blocks.}
\label{fig:csfi}
\end{figure}

\subsection{Cross-Scale Feature Integration}\label{sec:CSFI}
Our texture transformer can be further stacked in a cross-scale way with a cross-scale feature integration module. The architecture is shown in Figure~\ref{fig:csfi}. Stacked texture transformers output the synthesized features for three resolution scales ($1\times$, $2\times$ and $4\times$), such that the texture features of different scales can be fused into the LR image. To learn a better representation across different scales, inspired by~\cite{sun2019high, zeng2019learning}, we propose a cross-scale feature integration module (CSFI) to exchange information among the features of different scales. A CSFI module is applied each time the LR feature is up-sampled to the next scale. For the each scale inside the CSFI module, it receives the exchanged features from other scales by up/down-sampling, followed by a concatenation operation in the channel dimension. Then a convolutional layer will map the features into the original number of channels. In such a design, the texture features transferred from the stacked texture transformers are exchanged across each scale, which achieves a more powerful feature representation. This cross-scale feature integration module further improves the performance of our approach.

\subsection{Loss Function}\label{sec:loss}
There are 3 loss functions in our approach. The overall loss can be interpreted as:
\begin{align}
    \mathcal{L}_{overall} = \lambda_{rec} \mathcal{L}_{rec} + \lambda_{adv} \mathcal{L}_{adv} + \lambda_{per} \mathcal{L}_{per}.
\end{align}

\noindent\textbf{Reconstruction loss.} The first loss is the reconstruction loss:
\begin{align}
    \mathcal{L}_{rec} = \frac{1}{CHW} \left \| I^{HR} - I^{SR} \right \|_1,
\end{align}
where $(C, H, W)$ is the size of the HR. We utilize $L_1$ loss which has been demonstrated to be sharper for performance and easier for convergence compared to $L_2$ loss.

\noindent\textbf{Adversarial loss.} Generative adversarial networks~\cite{goodfellow2014generative} are proved effective in generating clear and visually favorable images. Here we adopt WGAN-GP~\cite{gulrajani2017improved}, which proposes a penalization of gradient norm to replace weight clipping, resulting in more stable training and better performance. This loss can be interpreted as:
\begin{align}
    \mathcal{L}_D &= \mathop{\mathbb{E}}\limits_{\tilde{x} \sim \mathbb{P}_g} \big[ D(\tilde{x}) \big] - \mathop{\mathbb{E}}\limits_{x \sim \mathbb{P}_r} \big[ D(x) \big] + \notag \\
    &\ \ \ \ \  \lambda \mathop{\mathbb{E}}\limits_{\hat{x} \sim \mathbb{P}_{\hat{x}}} \big[(\left \| \nabla_{\hat{x}} D(\hat{x}) \right \|_2 - 1)^2 \big], \\
    \mathcal{L}_G &= - \mathop{\mathbb{E}}\limits_{\tilde{x} \sim \mathbb{P}_g}\big[ D(\tilde{x}) \big].
\end{align}

\noindent\textbf{Perceptual loss.} Perceptual loss has been demonstrated useful to improve visual quality and has already been used in~\cite{johnson2016perceptual, ledig2017photo, sajjadi2017enhancenet, zhang2019image}. The key idea of perceptual loss is to enhance the similarity in feature space between the prediction image and the target image. Here our perceptual loss contains two parts:
\begin{align}
    \mathcal{L}_{per} =& \frac{1}{C_iH_iW_i} \left \| \phi^{vgg}_i(I^{SR}) - \phi^{vgg}_i(I^{HR})\right \|^2_2 + \notag \\
                      & \frac{1}{C_jH_jW_j} \left \| \phi^{lte}_j(I^{SR}) - T  \right\|^2_2,
\end{align}
where the first part is a traditional perceptual loss, in which $\phi^{vgg}_i(\cdot)$ denotes the $i$-th layer's feature map of VGG19, and $(C_i, H_i, W_i)$ represents the shape of the feature map at that layer. $I^{SR}$ is the predicted SR image. The second part in our perceptual loss is a transferal perceptual loss, in which $\phi^{lte}_j(\cdot)$ denotes the texture feature map extracted from the $j$-th layer of the proposed LTE, and $(C_j, H_j, W_j)$ represents that layer's shape. $T$ is the transferred HR texture features in Figure~\ref{fig:texture_transformer}. This transferal perceptual loss constraints the predicted SR image to have similar texture features to the transferred texture features $T$, which makes our approach to transfer the Ref textures more effectively.

\subsection{Implementation Details}\label{sec:implement}
The learnable texture extractor contains 5 convolutional layers and 2 pooling layers which outputs texture features in three different scales. To reduce the consumption of both time and GPU memory, the relevance embedding is only applied to the smallest scale and further propagated to other scales. For the discriminator, we adopt the same network used in SRNTT~\cite{zhang2019image} and remove all BN layers. During training, we augment the training images by randomly horizontally and vertically flipping followed by randomly rotating $90^{\circ}$, $180^{\circ}$ and $270^{\circ}$. Each mini-batch contains 9 LR patches with size $40 \times 40$ along with 9 HR and Ref patches with size $160 \times 160$. The weight coefficients for $\mathcal{L}_{rec}$, $\mathcal{L}_{adv}$ and $\mathcal{L}_{per}$ are 1, 1e-3 and 1e-2, respectively. Adam optimizer with  $\beta_1=0.9$, $\beta_2=0.999$, and $\epsilon=$1e-8 is used with learning rate of 1e-4. We first warm up the network for 2 epochs where only $\mathcal{L}_{rec}$ is applied. After that, all losses are involved to train another 50 epochs.

\section{Experiments}\label{sec:expr}

\subsection{Datasets and Metrics}
To evaluate our method, we train and test our model on the recently proposed RefSR dataset, CUFED5~\cite{zhang2019image}. The training set in CUFED5 contains 11,871 pairs, each pair consisting of an input image and a reference image. There are 126 testing images in CUFED5 testing set, each accompanied by 4 reference images with different similarity levels. In order to evaluate the generalization performance of TTSR trained on CUFED5, we additionally test TTSR on Sun80~\cite{sun2012super}, Urban100~\cite{huang2015single}, and Manga109~\cite{matsui2017sketch}. Sun80 contains 80 natural images, each paired with several reference images. For Urban100, we use the same setting as~\cite{zhang2019image} to regard its LR images as the reference images. Such a design enables an explicit process of self-similar searching and transferring since Urban100 are all building images with strong self-similarity. For Manga109 which also lacks the reference images, we randomly sample HR images in this dataset as the reference images. Since this dataset is constructed with lines, curves and flat colored regions which are all common patterns. Even with a randomly picked HR Ref image, our method can still utilize these common patterns and achieve good results. The SR results are evaluated on PSNR and SSIM on Y channel of YCbCr space.

\subsection{Evaluation}
To evaluate the effectiveness of TTSR, we compare our model with other state-of-the-art SISR and RefSR methods. The SISR methods include SRCNN~\cite{dong2015image}, MDSR~\cite{lim2017enhanced}, RDN~\cite{zhang2018residual}, RCAN~\cite{zhang2018image}, SRGAN~\cite{ledig2017photo}, ENet~\cite{sajjadi2017enhancenet}, ESRGAN~\cite{wang2018esrgan}, RSRGAN~\cite{zhang2019ranksrgan}, among which RCAN has achieved state-of-the-art performance on both PSNR and SSIM in recent years. RSRGAN is considered to achieve the state-of-the-art visual quality. As for RefSR methods, CrossNet~\cite{zheng2018crossnet} and SRNTT~\cite{zhang2019image} are two state-of-the-art methods recently, which significantly outperform previous RefSR methods. All experiments are performed with a scaling factor of $4\times$ between LR and HR images.

\begin{table}
\caption{PSNR/SSIM comparison among different SR methods on four different datasets. Methods are grouped by SISR methods (top) and RefSR methods (down). Red numbers denote the highest scores while blue numbers denote the second highest scores.}
\label{tab:quantitative_refsr}
\centering
\scalebox{0.74}{
\begin{tabular}{l|cccc}
\hline
Method & CUFED5 & Sun80 & Urban100 & Manga109 \\
\hline
SRCNN~\cite{dong2015image} & 25.33 / .745 & 28.26 / .781 & 24.41 / .738 &27.12 / .850\\
MDSR~\cite{lim2017enhanced} & 25.93 / .777 & 28.52 / .792 & {\color{blue}25.51} / .783 & 28.93 / .891 \\
RDN~\cite{zhang2018residual} &25.95 / .769 &29.63 / .806 &25.38 / .768 & 29.24 / .894\\
RCAN~\cite{zhang2018image} &26.06 / .769 &{\color{blue}29.86} / {\color{blue}.810} &25.42 / .768 &{\color{blue}29.38} / {\color{blue}.895}\\
SRGAN~\cite{ledig2017photo} & 24.40 / .702 & 26.76 / .725 & 24.07 / .729 & 25.12 / .802\\
ENet~\cite{sajjadi2017enhancenet} & 24.24 / .695 & 26.24 / .702 & 23.63 / .711 & 25.25 / .802\\  
ESRGAN~\cite{wang2018esrgan} &21.90 / .633  &24.18 / .651 &20.91 / .620 & 23.53 / .797\\
RSRGAN~\cite{zhang2019ranksrgan} & 22.31 / .635 & 25.60 / .667 & 21.47 / .624 & 25.04 / .803\\
\hline
CrossNet~\cite{zheng2018crossnet} & 25.48 / .764 & 28.52 / .793 & 25.11 / .764 & 23.36 / .741\\
SRNTT-\textit{rec}~\cite{zhang2019image} & {\color{blue}26.24} / {\color{blue}.784} & 28.54 / .793 & 25.50 / {\color{blue}.783} &28.95 / .885\\
SRNTT~\cite{zhang2019image} &25.61 / .764 &27.59 / .756 &25.09 / .774 &27.54 / .862\\
TTSR-\textit{rec}  & {\color{red}27.09} / {\color{red}.804} & {\color{red}30.02} / {\color{red}.814} & {\color{red}25.87} / {\color{red}.784}  & {\color{red}30.09} / {\color{red}.907}\\
TTSR &25.53 / .765 &28.59 / .774 &24.62 / .747 & 28.70 / .886\\
\hline
\end{tabular}
}
\vspace{-0.4cm}
\end{table}

\noindent\textbf{Quantitative Evaluation.} For fair comparison, we follow the setting in SRNTT~\cite{zhang2019image} to train all the methods on CUFED5 training set, and test on CUFED5 testing set, Sun80, Urban100 and Manga109 datasets. For SR methods, there is a fact that training with adversarial loss usually achieves better visual quality but shrinks the number of PSNR and SSIM. Therefore, we train another version of our model which is optimized only on reconstruction loss named TTSR-\textit{rec} for fair comparison on PSNR and SSIM.

Table~\ref{tab:quantitative_refsr} shows the quantitative evaluation results. Red numbers denote the highest scores while blue numbers denote the second highest scores. As shown in the comparison results, TTSR-\textit{rec} significantly outperforms both state-of-the-art SISR methods and state-of-the-art RefSR methods on all four testing datasets. Among the methods which aim to achieve better visual quality with adversarial loss, our model still has the best performance on Sun80 and Manga109 datasets. On the other two datasets, CUFED5 and Urban100, our model achieves comparable performance with the state-of-the-art models. The quantitative comparison results demonstrate the superiority of our proposed TTSR over state-of-the-art SR approaches.

\begin{figure*}[t]
    \centering
	\begin{tabular}{p{0.203\linewidth}<{\centering}p{0.202\linewidth}<{\centering}p{0.202\linewidth}<{\centering}p{0.202\linewidth}<{\centering}}
		Ground-truth & RDN~\cite{zhang2018residual} & RCAN~\cite{zhang2018image} & RSRGAN~\cite{zhang2019ranksrgan} \\
		\hline
		Reference & CrossNet~\cite{zheng2018crossnet} & SRNTT~\cite{zhang2019image} & TTSR (Ours)\\
	\end{tabular}
    \includegraphics[width=0.9\linewidth, page=2]{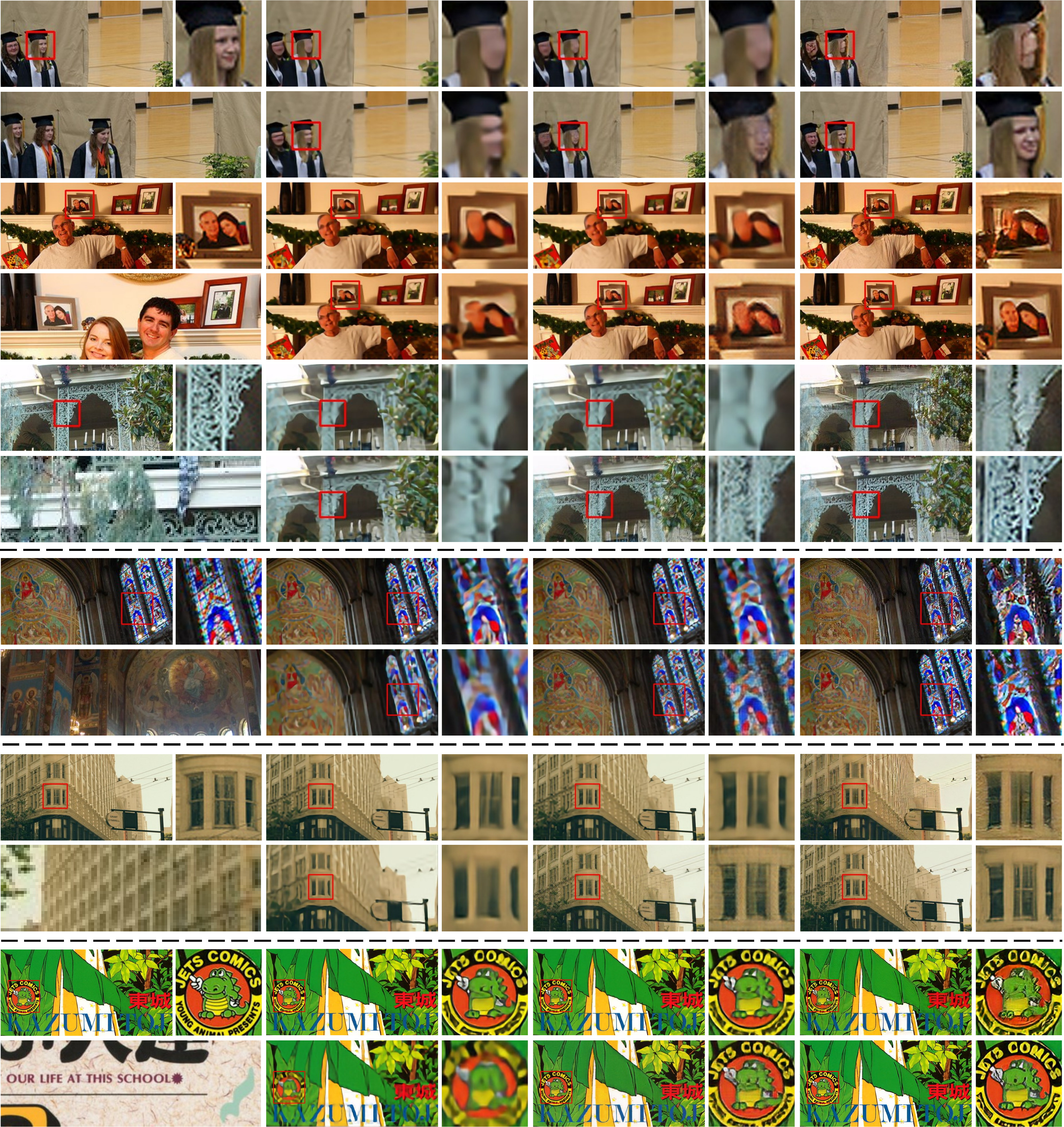}
    \caption{Visual comparison among different SR methods on CUFED5 testing set (top three examples), Sun80~\cite{sun2012super} (the forth example), Urban100~\cite{huang2015single} (the fifth example), and Manga109~\cite{matsui2017sketch} (the last example).}
    \label{fig:refsr}
    \vspace{-0.3cm}
\end{figure*}

\noindent\textbf{Qualitative Evaluation.} Our model also achieves the best performance on visual quality as shown in Figure~\ref{fig:refsr}. TTSR can transfer more accurate HR textures from the reference image to generate favorable results, as shown in the first three examples in Figure~\ref{fig:refsr}. Even if the reference image is not that globally relevant to the input image, our TTSR can still extract finer textures from local regions and transfer effective textures into the predicted SR result, as shown in the last three examples in Figure~\ref{fig:refsr}.

\begin{figure}[t]
    \centering
    \includegraphics[width=\linewidth]{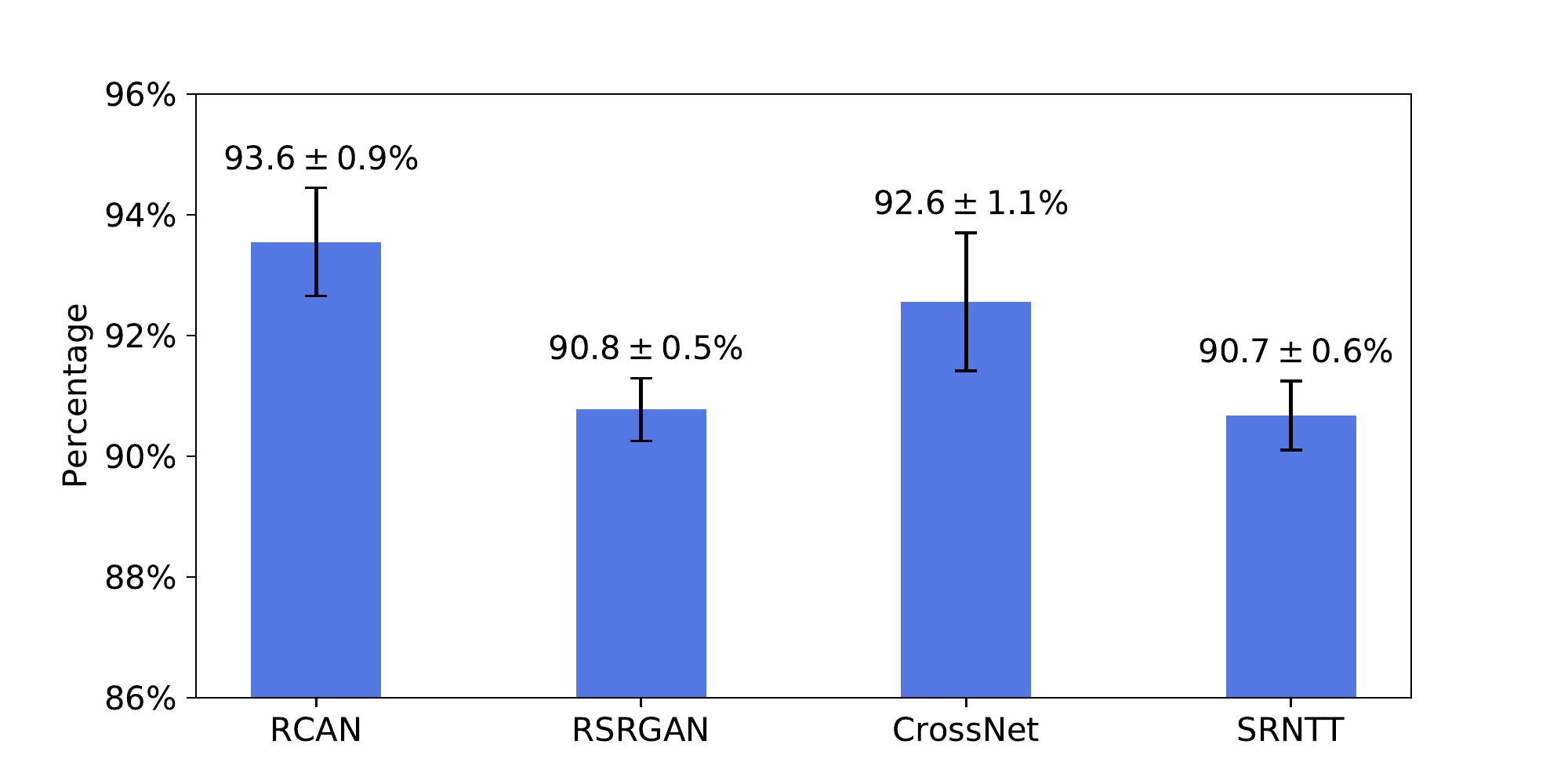}
    \caption{User study results. Values on Y-axis indicate the percentage of users that prefer TTSR over other approaches.}
    \label{fig:user_study}
\end{figure}

To further verify the superior visual quality of our approach, we conduct a user study where TTSR is compared with four SOTA approaches, including RCAN~\cite{zhang2018image}, RSRGAN~\cite{zhang2019ranksrgan}, CrossNet~\cite{zheng2018crossnet} and SRNTT~\cite{zhang2019image}. There are 10 subjects involved in this user study and 2,520 votes are collected on the CUFED5 testing set. For each comparison process, we provide the users with two images which include one TTSR image. Users are asked to select the one with higher visual quality. Figure~\ref{fig:user_study} shows the results of our user study, where the values on Y-axis represent the percentage of users that prefer TTSR over other approaches. As we can see, the proposed TTSR significantly outperforms other approaches with over $90\%$ of users voting for ours, which verifies the favorable visual quality of TTSR.

\subsection{Ablation Study}
In this section, we verify the effectiveness of different modules in our approach, including the texture transformer, the cross-scale feature integration, the adversarial loss and the transferal perceptual loss. In addition, we also discuss the influence of different reference similarity on TTSR.

\begin{table}[t]
\caption{Ablation study on texture transformer.}
\label{tab:ablation_TT}
\begin{center}
\scalebox{0.95}{
\begin{tabular}{l|c|c|c|c}
\hline
Method &HA &SA & LTE & PSNR/SSIM\\
\hline
Base            &            &            &            &26.34 / .780\\
Base+HA         & \checkmark &            &            &26.59 / .786\\
Base+HA+SA      & \checkmark & \checkmark &            &26.81 / .795\\
Base+HA+SA+LTE  & \checkmark & \checkmark & \checkmark &26.92 / .797\\
\hline
\end{tabular}
}
\end{center}
\vspace{-0.3cm}
\end{table}

\begin{table}[t]
\caption{Ablation study on CSFI.}
\label{tab:ablation_CSFI}
\begin{center}
\scalebox{0.9}{
\begin{tabular}{l|c|c|c|c}
\hline
Method &CSFI &numC & param. & PSNR/SSIM\\
\hline
Base+TT          &            & 64 &4.42M &26.92 / .797 \\
Base+TT+CSFI     & \checkmark & 64 &6.42M &27.09 / .804 \\
Base+TT(C80)     &            & 80 &6.53M &26.93 / .797 \\
Base+TT(C96)     &            & 96 &9.10M &26.98 / .799 \\
\hline
\end{tabular}
}
\end{center}
\vspace{-0.6cm}
\end{table}

\noindent\textbf{Texture transformer.} \label{sec:ablation_TT}
Our texture transformer contains mainly four parts: the learnable texture extractor (LTE), the relevance embedding module, the hard-attention module for feature transfer (HA) and the soft-attention module for feature synthesis (SA). Ablation results are shown in Table~\ref{tab:ablation_TT}. We re-implement SRNTT~\cite{zhang2019image} as our ``Base'' model by only removing all BN layers and Ref part. On top of the baseline model, we progressively add HA, SA, and LTE. Models without LTE use the VGG19 features to do relevance embedding. As we can see, when HA is added, the PSNR performance can be improved from $26.34$ to $26.59$, which verifies the effectiveness of the hard-attention module for feature transfer. When SA is involved, relevant texture features will be enhanced while the less relevant ones will be relieved during the feature synthesizing. This further boosts the performance to $26.81$. When replacing VGG with the proposed LTE, the PSNR is finally increased to $26.92$, which proves the superiority of joint feature embedding in LTE.

\begin{figure}[t]
    \centering
    \includegraphics[width=\linewidth, page=3]{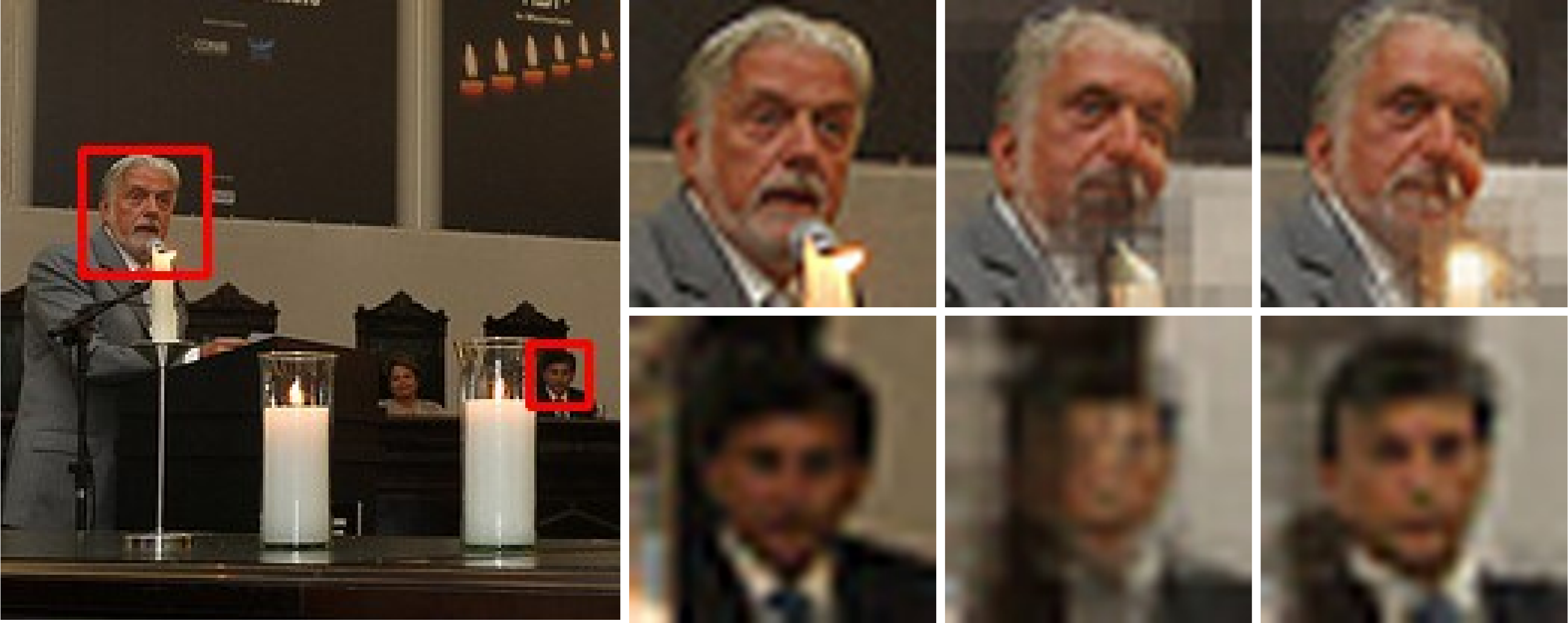}
    \begin{tabular}{p{0.35\linewidth}<{\centering}p{0.14\linewidth}<{\centering}p{0.16\linewidth}<{\centering}p{0.15\linewidth}<{\centering}}
		\small{GT }& \small{GT} & \small{SRNTT} & \small{TTSR(ours)} \\
	\end{tabular}
    \caption{Comparison of the transferred original images between SRNTT and TTSR.}
    \label{fig:swap_vgg_dte}
\end{figure}

To further verify the effectiveness of our LTE, we use the hard attention map to transfer the original image. It is expected that a better feature representation can transfer more accurate textures from the original images. Figure~\ref{fig:swap_vgg_dte} shows the transferred original image by VGG19 in SRNTT and LTE in TTSR. In this figure, TTSR can transfer more accurate reference textures and generate a globally favorable result, which further proves the effectiveness of our LTE.

\noindent\textbf{Cross-scale feature integration.} \label{sec:ablation_CSFI}
On top of the texture transformer, CSFI can further enable texture recovery from different resolution scales ($1\times$, $2\times$ and $4\times$). We conduct an ablation study in Table~\ref{tab:ablation_CSFI}. The first row shows the performance of our model with only TT, while the second row proves the effectiveness of CSFI, which brings $0.17$ increase on PSNR metric. In order to verify that the performance improvement is not brought by the increase of parameter size, we increase the channel number of ``Base+TT'' model to 80 and 96. As we can see, there is almost no growth of  ``Base+TT(C80)'' which has almost the same parameter number as ``Base+TT+CSFI''. Even if we increase the parameter number to $9.10$M to obtain ``Base+TT(C96)'' model, there is still a performance gap. This demonstrates that CSFI can efficiently utilize the reference texture information with a relatively smaller parameter size.

\begin{figure}[t]
\centering
\includegraphics[width=\linewidth, page=4]{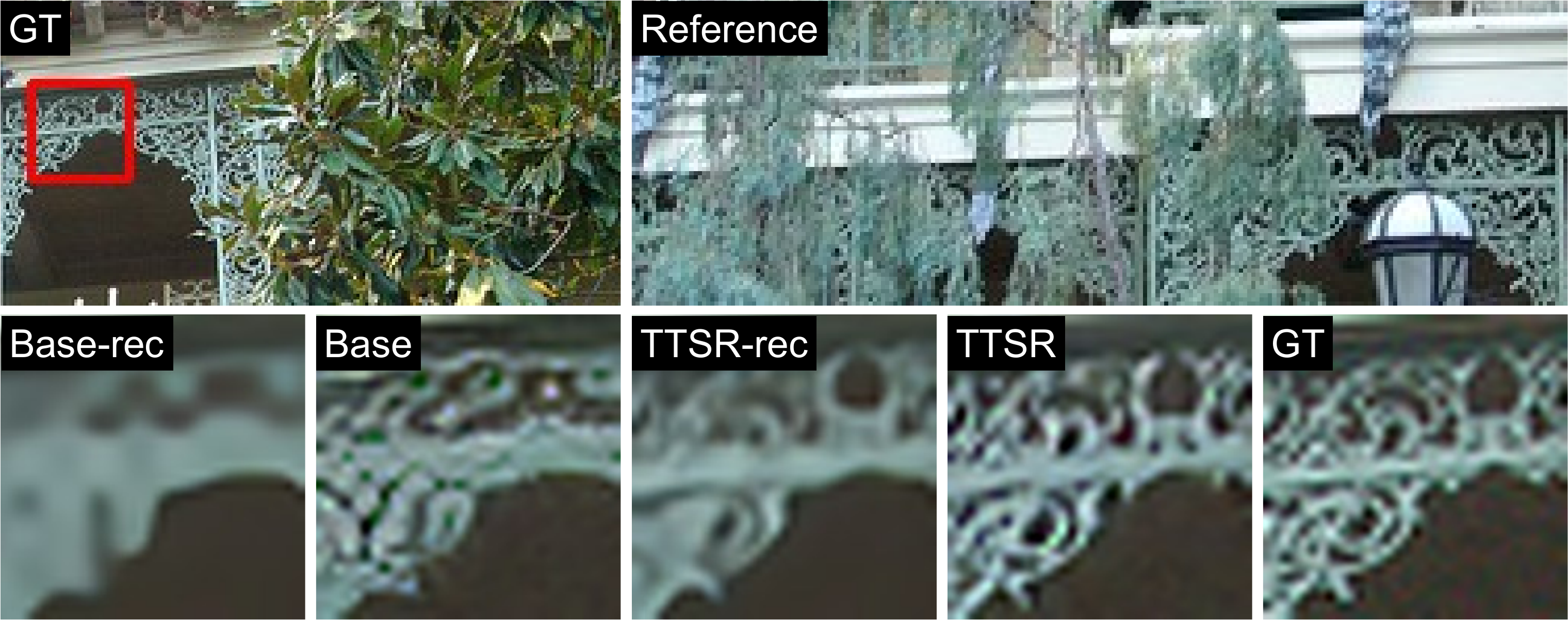}
\caption{Qualitative comparison on ``Base(-\textit{rec})'' and TTSR(-\textit{rec}) (TTSR can be interpreted as ``Base+TT(HA+SA+LTE)+CSFI'').}
\label{fig:gan}
\vspace{-0.3cm}
\end{figure}

\noindent\textbf{Adversarial loss.} \label{sec:ablation_GAN}
To make sure that the improvement of perceptual quality benefits from model design rather than the adversarial loss. We conduct an ablation among ``Base-\textit{rec}'', ``Base'', TTSR-\textit{rec} and TTSR, where TTSR can be interpreted as ``Base+TT+CSFI'' and ``-\textit{rec}'' indicates training with only reconstruction loss. Figure~\ref{fig:gan} shows that even if without the perceptual and adversarial loss, TTSR-\textit{rec} can still utilize the Ref image and recover more details than ``Base-\textit{rec}''. With all losses enabled, TTSR achieves the best visual result.

\begin{figure}[t]
    \centering
    \includegraphics[width=0.63\linewidth, page=5]{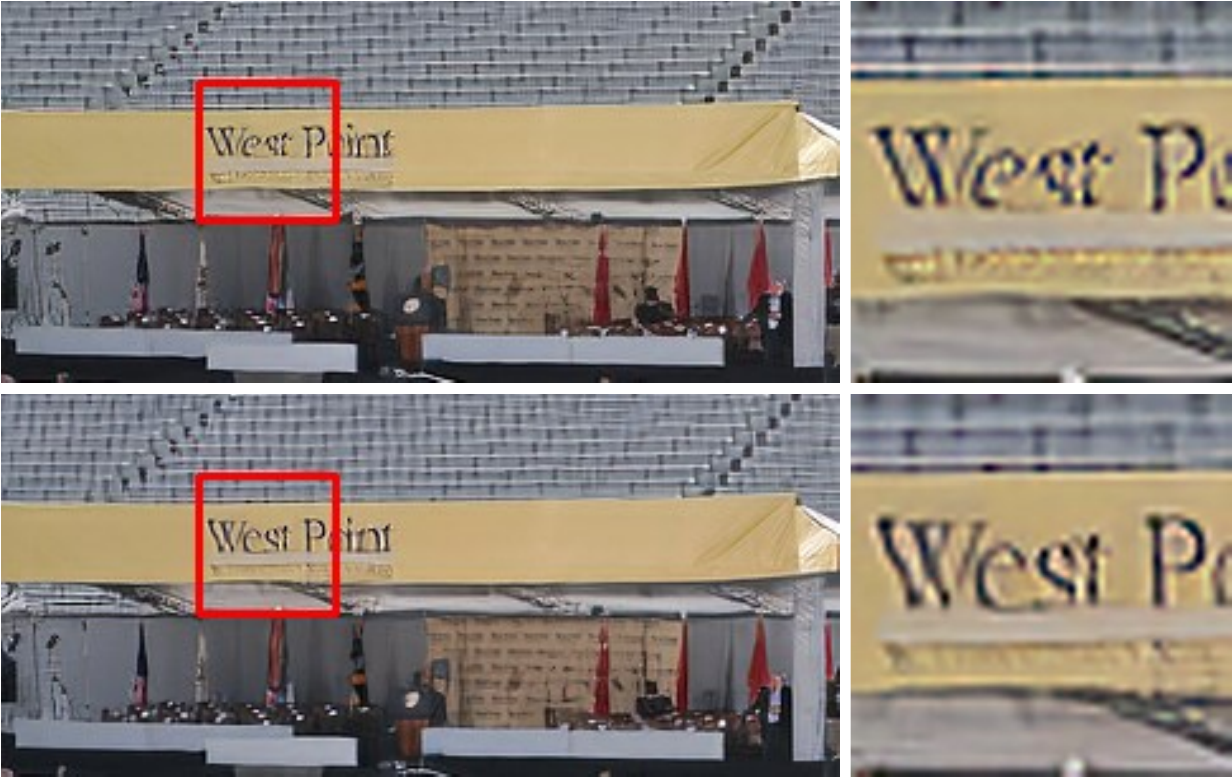}
    \caption{Comparison between TTSR trained without (top) and with (bottom) transferal perceptual loss.}
    \label{fig:ablation_tploss}
\end{figure}

\noindent\textbf{Transferal perceptual loss.} \label{sec:ablation_TPL}
The transferal perceptual loss constraints the LTE's features between the predicted SR image and the transferred image $T$ to be similar. As shown in Figure~\ref{fig:ablation_tploss}, using this loss is able to transfer textures in a more effective way which achieves visually pleasing results. In addition, this loss also improves the quantitative metrics PSNR and SSIM of TTSR from $25.20 / .757$ to $25.53 / .765$.

\noindent\textbf{Influence of different reference similarity.} \label{sec:ablation_DL}
To study how relevance between LR and Ref images influences the results of TTSR, we conduct experiments on CUFED5 testing set, which has reference images of different relevance levels. Table~\ref{tab:ablation_rrl} shows the results of five relevance levels, in which ``L1'' to ``L4'' represent the reference images provided by CUFED5 testing set where L1 is the most relevant level while L4 is the least relevant one. ``LR'' means using the input image itself as the reference image. As shown in Table~\ref{tab:ablation_rrl}, TTSR using L1 as the reference image achieves the best performance. When using LR as the reference image, TTSR still performs better than previous state-of-the-art RefSR approaches.

\begin{table}
\caption{Ablation study on reference images of different similarity.}
\label{tab:ablation_rrl}
\begin{center}
\begin{tabular}{l|c|c|c}
\hline
Level &CrossNet &SRNTT-\textit{rec} &TTSR-\textit{rec}\\
\hline
L1 &25.48 / .764 &26.15 / .781 &26.99 / .800 \\
L2 &25.48 / .764 &26.04 / .776 &26.74 / .791 \\
L3 &25.47 / .763 &25.98 / .775 &26.64 / .788 \\
L4 &25.46 / .763 &25.95 / .774 &26.58 / .787 \\
LR &25.46 / .763 &25.91 / .776 &26.43 / .782 \\
\hline
\end{tabular}
\vspace{-0.3cm}
\end{center}
\end{table}

\section{Conclusion}\label{sec:conclusion}
In this paper, we propose a novel \textbf{T}exture \textbf{T}ransformer Network for Image \textbf{S}uper-\textbf{R}esolution (TTSR) which transfers HR textures from the Ref to LR image. The proposed texture transformer consists of a learnable texture extractor which learns a jointly feature embedding for further attention computation and two attention based modules which transfer HR textures from the Ref image. Furthermore, the proposed texture transformer can be stacked in a cross-scale way with the proposed CSFI module to learn a more powerful feature representation. Extensive experiments demonstrate the superior performance of our TTSR over state-of-the-art approaches on both quantitative and qualitative evaluations. In the future, we will further extend the proposed texture transformer to general image generation tasks.

\textbf{Acknowledgement} This paper is partially supported by NSFC (No. 61772330, 61533012, 61876109), the pre-research project (No. 61403120201), Shanghai Key Laboratory of Crime Scene Evidence (2017XCWZK01) and the Interdisciplinary Program of Shanghai Jiao Tong University (YG2019QNA09).

{\small
\bibliographystyle{ieee_fullname}
\bibliography{ttsr}

\begin{thebibliography}{10}\itemsep=-1pt

\bibitem{boominathan2014improving}
Vivek Boominathan, Kaushik Mitra, and Ashok Veeraraghavan.
\newblock Improving resolution and depth-of-field of light field cameras using
  a hybrid imaging system.
\newblock In {\em ICCP}, pages 1--10, 2014.

\bibitem{dai2019second}
Tao Dai, Jianrui Cai, Yongbing Zhang, Shu-Tao Xia, and Lei Zhang.
\newblock Second-order attention network for single image super-resolution.
\newblock In {\em CVPR}, pages 11065--11074, 2019.

\bibitem{dong2015image}
Chao Dong, Chen~Change Loy, Kaiming He, and Xiaoou Tang.
\newblock Image super-resolution using deep convolutional networks.
\newblock {\em TPAMI}, 38(2):295--307, 2015.

\bibitem{dong2016accelerating}
Chao Dong, Chen~Change Loy, and Xiaoou Tang.
\newblock Accelerating the super-resolution convolutional neural network.
\newblock In {\em ECCV}, pages 391--407, 2016.

\bibitem{freedman2011image}
Gilad Freedman and Raanan Fattal.
\newblock Image and video upscaling from local self-examples.
\newblock {\em ACM Transactions on Graphics (TOG)}, 30(2):1--11, 2011.

\bibitem{freeman2002example}
William~T Freeman, Thouis~R Jones, and Egon~C Pasztor.
\newblock Example-based super-resolution.
\newblock {\em IEEE Computer graphics and Applications}, 22(2):56--65, 2002.

\bibitem{goodfellow2014generative}
Ian Goodfellow, Jean Pouget-Abadie, Mehdi Mirza, Bing Xu, David Warde-Farley,
  Sherjil Ozair, Aaron Courville, and Yoshua Bengio.
\newblock Generative adversarial nets.
\newblock In {\em NeurIPS}, pages 2672--2680, 2014.

\bibitem{gulrajani2017improved}
Ishaan Gulrajani, Faruk Ahmed, Martin Arjovsky, Vincent Dumoulin, and Aaron~C
  Courville.
\newblock Improved training of wasserstein gans.
\newblock In {\em NeurIPS}, pages 5767--5777, 2017.

\bibitem{he2016deep}
Kaiming He, Xiangyu Zhang, Shaoqing Ren, and Jian Sun.
\newblock Deep residual learning for image recognition.
\newblock In {\em CVPR}, pages 770--778, 2016.

\bibitem{huang2017densely}
Gao Huang, Zhuang Liu, Laurens Van Der~Maaten, and Kilian~Q Weinberger.
\newblock Densely connected convolutional networks.
\newblock In {\em CVPR}, pages 4700--4708, 2017.

\bibitem{huang2015single}
Jia-Bin Huang, Abhishek Singh, and Narendra Ahuja.
\newblock Single image super-resolution from transformed self-exemplars.
\newblock In {\em CVPR}, pages 5197--5206, 2015.

\bibitem{IraniImproving}
Michal Irani and Shmuel Peleg.
\newblock Improving resolution by image registration.
\newblock {\em CVGIP}, 53(3):231--239, 1991.

\bibitem{johnson2016perceptual}
Justin Johnson, Alexandre Alahi, and Li Fei-Fei.
\newblock Perceptual losses for real-time style transfer and super-resolution.
\newblock In {\em ECCV}, pages 694--711, 2016.

\bibitem{kim2016accurate}
Jiwon Kim, Jung Kwon~Lee, and Kyoung Mu~Lee.
\newblock Accurate image super-resolution using very deep convolutional
  networks.
\newblock In {\em CVPR}, pages 1646--1654, 2016.

\bibitem{kim2016deeply}
Jiwon Kim, Jung Kwon~Lee, and Kyoung Mu~Lee.
\newblock Deeply-recursive convolutional network for image super-resolution.
\newblock In {\em CVPR}, pages 1637--1645, 2016.

\bibitem{ledig2017photo}
Christian Ledig, Lucas Theis, Ferenc Husz{\'a}r, Jose Caballero, Andrew
  Cunningham, Alejandro Acosta, Andrew Aitken, Alykhan Tejani, Johannes Totz,
  Zehan Wang, et~al.
\newblock Photo-realistic single image super-resolution using a generative
  adversarial network.
\newblock In {\em CVPR}, pages 4681--4690, 2017.

\bibitem{lim2017enhanced}
Bee Lim, Sanghyun Son, Heewon Kim, Seungjun Nah, and Kyoung Mu~Lee.
\newblock Enhanced deep residual networks for single image super-resolution.
\newblock In {\em CVPR Workshops}, pages 136--144, 2017.

\bibitem{liu2018non}
Ding Liu, Bihan Wen, Yuchen Fan, Chen~Change Loy, and Thomas~S Huang.
\newblock Non-local recurrent network for image restoration.
\newblock In {\em NeurIPS}, pages 1673--1682, 2018.

\bibitem{ma2018gan}
Shuang Ma, Jianlong Fu, Chang Wen~Chen, and Tao Mei.
\newblock Da-gan: Instance-level image translation by deep attention generative
  adversarial networks.
\newblock In {\em CVPR}, pages 5657--5666, 2018.

\bibitem{matsui2017sketch}
Yusuke Matsui, Kota Ito, Yuji Aramaki, Azuma Fujimoto, Toru Ogawa, Toshihiko
  Yamasaki, and Kiyoharu Aizawa.
\newblock Sketch-based manga retrieval using manga109 dataset.
\newblock {\em Multimedia Tools and Applications}, 76(20):21811--21838, 2017.

\bibitem{oktay2016multi}
Ozan Oktay, Wenjia Bai, Matthew Lee, Ricardo Guerrero, Konstantinos Kamnitsas,
  Jose Caballero, Antonio de Marvao, Stuart Cook, Declan O’Regan, and Daniel
  Rueckert.
\newblock Multi-input cardiac image super-resolution using convolutional neural
  networks.
\newblock In {\em MICCAI}, pages 246--254, 2016.

\bibitem{sajjadi2017enhancenet}
Mehdi~SM Sajjadi, Bernhard Scholkopf, and Michael Hirsch.
\newblock Enhancenet: Single image super-resolution through automated texture
  synthesis.
\newblock In {\em ICCV}, pages 4491--4500, 2017.

\bibitem{shi2016real}
Wenzhe Shi, Jose Caballero, Ferenc Husz{\'a}r, Johannes Totz, Andrew~P Aitken,
  Rob Bishop, Daniel Rueckert, and Zehan Wang.
\newblock Real-time single image and video super-resolution using an efficient
  sub-pixel convolutional neural network.
\newblock In {\em CVPR}, pages 1874--1883, 2016.

\bibitem{simonyan2014very}
Karen Simonyan and Andrew Zisserman.
\newblock Very deep convolutional networks for large-scale image recognition.
\newblock {\em arXiv preprint arXiv:1409.1556}, 2014.

\bibitem{sun2019high}
Ke Sun, Yang Zhao, Borui Jiang, Tianheng Cheng, Bin Xiao, Dong Liu, Yadong Mu,
  Xinggang Wang, Wenyu Liu, and Jingdong Wang.
\newblock High-resolution representations for labeling pixels and regions.
\newblock {\em arXiv preprint arXiv:1904.04514}, 2019.

\bibitem{sun2012super}
Libin Sun and James Hays.
\newblock Super-resolution from internet-scale scene matching.
\newblock In {\em ICCP}, pages 1--12, 2012.

\bibitem{tai2017image}
Ying Tai, Jian Yang, and Xiaoming Liu.
\newblock Image super-resolution via deep recursive residual network.
\newblock In {\em CVPR}, pages 3147--3155, 2017.

\bibitem{tai2017memnet}
Ying Tai, Jian Yang, Xiaoming Liu, and Chunyan Xu.
\newblock Memnet: A persistent memory network for image restoration.
\newblock In {\em ICCV}, pages 4539--4547, 2017.

\bibitem{timofte2013anchored}
Radu Timofte, Vincent De~Smet, and Luc Van~Gool.
\newblock Anchored neighborhood regression for fast example-based
  super-resolution.
\newblock In {\em ICCV}, pages 1920--1927, 2013.

\bibitem{tong2017image}
Tong Tong, Gen Li, Xiejie Liu, and Qinquan Gao.
\newblock Image super-resolution using dense skip connections.
\newblock In {\em ICCV}, pages 4799--4807, 2017.

\bibitem{vaswani2017attention}
Ashish Vaswani, Noam Shazeer, Niki Parmar, Jakob Uszkoreit, Llion Jones,
  Aidan~N Gomez, {\L}ukasz Kaiser, and Illia Polosukhin.
\newblock Attention is all you need.
\newblock In {\em NeurIPS}, pages 5998--6008, 2017.

\bibitem{wang2018esrgan}
Xintao Wang, Ke Yu, Shixiang Wu, Jinjin Gu, Yihao Liu, Chao Dong, Yu Qiao, and
  Chen Change~Loy.
\newblock Esrgan: Enhanced super-resolution generative adversarial networks.
\newblock In {\em ECCV Workshops}, 2018.

\bibitem{wang2016light}
Yuwang Wang, Yebin Liu, Wolfgang Heidrich, and Qionghai Dai.
\newblock The light field attachment: Turning a dslr into a light field camera
  using a low budget camera ring.
\newblock {\em IEEE TVCG}, 23(10):2357--2364, 2016.

\bibitem{xu2018attngan}
Tao Xu, Pengchuan Zhang, Qiuyuan Huang, Han Zhang, Zhe Gan, Xiaolei Huang, and
  Xiaodong He.
\newblock Attngan: Fine-grained text to image generation with attentional
  generative adversarial networks.
\newblock In {\em CVPR}, pages 1316--1324, 2018.

\bibitem{yildirim2012novel}
Deniz Y{\i}ld{\i}r{\i}m and O{\u{g}}uz G{\"u}ng{\"o}r.
\newblock A novel image fusion method using ikonos satellite images.
\newblock {\em Journal of Geodesy and Geoinformation}, 1(1):75--83, 2012.

\bibitem{yue2013landmark}
Huanjing Yue, Xiaoyan Sun, Jingyu Yang, and Feng Wu.
\newblock Landmark image super-resolution by retrieving web images.
\newblock {\em IEEE TIP}, 22(12):4865--4878, 2013.

\bibitem{zeng2019learning}
Yanhong Zeng, Jianlong Fu, Hongyang Chao, and Baining Guo.
\newblock Learning pyramid-context encoder network for high-quality image
  inpainting.
\newblock In {\em CVPR}, pages 1486--1494, 2019.

\bibitem{zhang2019ranksrgan}
Wenlong Zhang, Yihao Liu, Chao Dong, and Yu Qiao.
\newblock Ranksrgan: Generative adversarial networks with ranker for image
  super-resolution.
\newblock In {\em ICCV}, pages 3096--3105, 2019.

\bibitem{zhang2018image}
Yulun Zhang, Kunpeng Li, Kai Li, Lichen Wang, Bineng Zhong, and Yun Fu.
\newblock Image super-resolution using very deep residual channel attention
  networks.
\newblock In {\em ECCV}, pages 286--301, 2018.

\bibitem{zhang2018residual}
Yulun Zhang, Yapeng Tian, Yu Kong, Bineng Zhong, and Yun Fu.
\newblock Residual dense network for image super-resolution.
\newblock In {\em CVPR}, pages 2472--2481, 2018.

\bibitem{zhang2019image}
Zhifei Zhang, Zhaowen Wang, Zhe Lin, and Hairong Qi.
\newblock Image super-resolution by neural texture transfer.
\newblock In {\em CVPR}, pages 7982--7991, 2019.

\bibitem{zheng2017learning}
Haitian Zheng, Mengqi Ji, Lei Han, Ziwei Xu, Haoqian Wang, Yebin Liu, and Lu
  Fang.
\newblock Learning cross-scale correspondence and patch-based synthesis for
  reference-based super-resolution.
\newblock In {\em BMVC}, 2017.

\bibitem{zheng2018crossnet}
Haitian Zheng, Mengqi Ji, Haoqian Wang, Yebin Liu, and Lu Fang.
\newblock Crossnet: An end-to-end reference-based super resolution network
  using cross-scale warping.
\newblock In {\em ECCV}, pages 88--104, 2018.

\end{thebibliography}
}

\clearpage
\appendix
\renewcommand{\thesection}{\Alph{section}}
\renewcommand\thefigure{\thesection.\arabic{figure}}
\renewcommand\thetable{\thesection.\arabic{table}}

\setcounter{section}{0}
\section*{Supplementary}
In this supplementary material, Section~\ref{sec:network_structure} illustrates the details of TTSR's network structure. Section~\ref{sec:num_TT} provides additional analyses about the texture transformers on different scales. Section~\ref{sec:time} describes the comparison of the running time and the parameter number. Finally, more visual comparison results will be shown in Section~\ref{sec:visual}.

\section{Details of Network Structure}\label{sec:network_structure}
\setcounter{table}{0}
In this section, we will illustrate the detailed network structure of our approach TTSR, including the learnable texture extractor in the texture transformer, the generator with three stacked texture transformers and the discriminator. The structure of the learable texture extractor is shown in Table~\ref{tab:lte}, in which the layers \$0, \$3 and \$6 are used to search and transfer texture features in the texture transformer. Table~\ref{tab:generator} shows the details of the generator, and Table~\ref{tab:discriminator} illustrates the discriminator.

\begin{table}[htb]
\caption{Network structure of the learnable texture extractor. Conv($N_{in}$, $N_{out}$) indicates the convolutional layer with $N_{in}$ input channels and $N_{out}$ output channels. The kernel size is $3\times3$ for all convolutional layers. Pool($2\times2$) is the $2\times2$ pooling layer with stride 2.}
\vspace{-0.2cm}
\label{tab:lte}
\begin{center}
\begin{tabular}{|c|c|c|}
\hline
Id  &Layer Name \\
\hline
0 &Conv(3,64), ReLU \\
\hline
1 &Conv(64,64), ReLU \\
\hline
2 &Pool($2\times2$) \\
\hline
3 &Conv(64,128), ReLU \\
\hline
4 &Conv(128,128), ReLU \\
\hline
5 &Pool($2\times2$) \\
\hline
6 &Conv(128, 256), ReLU \\
\hline
\end{tabular}
\end{center}
\vspace{-0.4cm}
\end{table}

\begin{table*}[htb]
\caption{Network structure of the generator. Conv($N_{in}$, $N_{out}$) indicates the convolutional layer with $N_{in}$ input channels and $N_{out}$ output channels. The kernel size is $3\times3$ for all convolutional layers except that the last convolution uses $1\times1$ kernel. RB denotes the residual block without batch normalization layers and the ReLU layer after the skip connection. TT represents the texture transformer and PS is the $2\times$ pixel shuffle layer. $\uparrow$ indicates bicubic up-sampling followed by a $1\times1$ convolution, while $\downarrow$ denotes the strided convolution.}
\label{tab:generator}
\begin{center}
\scalebox{0.9}{
\begin{tabular}{|c|c|c|c|c|c|c|}
\hline
  &Id &Layer Name (scale$1\times$) &Id &Layer Name (scale$2\times$) &Id &Layer Name (scale$4\times$) \\
\hline
\multirow{4}*{Stage0} &1-0 &Conv(3,64), ReLU  & & & &\\
         \cline{2-3}
         &1-1 &RB $\times16$ & & & & \\
         \cline{2-3}
         &1-2 &Conv(64,64) & & & &\\
         \cline{2-3}
         &1-3 &\$1-0 + \$1-2 & & & &\\
\cline{1-3}
\multirow{4}*{Stage1}  &1-4 &TT  & & & &\\
        \cline{2-3}
        &1-5 &RB$\times 16$ & & & &\\
        \cline{2-3}
        &1-6 &Conv(64,64) & & & &\\
        \cline{2-3}
        &1-7 &\$1-4 + \$1-6 & & & &\\ 
\cline{1-5}
\multirow{7}*{Stage2}  &  & &2-0 &Conv(64,256), PS, ReLU(\$1-7) & &\\
        \cline{4-5}
        & & &2-1 &TT & &\\
        \cline{2-5}
        &1-8 &Concat(\$1-7 $||$ \$2-1$\downarrow$) &2-2 &Concat(\$1-7$\uparrow$ $||$ \$2-1) & &\\
        \cline{2-5}
        &1-9 &Conv(128,64), ReLU &2-3 &Conv(128,64), ReLU & &\\
        \cline{2-5}
        &1-10 &RB$\times8$ &2-4 &RB$\times8$ & &\\
        \cline{2-5}
        &1-11 &Conv(64,64) &2-5 &Conv(64,64) & &\\
        \cline{2-5}
        &1-12 &\$1-7 + \$1-11 &2-6 &\$2-1 + \$2-5 & &\\
\hline
\multirow{7}*{Stage3}  & & & & &4-0 &Conv(64,256), PS, ReLU(\$2-6)\\
        \cline{6-7}
        & & & & &4-1 &TT\\
        \cline{2-7}
        &1-13 &Concat(\$1-12 $||$ \$2-6$\downarrow$ $||$ \$4-1$\downarrow$) &2-7 &Concat(\$1-12$\uparrow$ $||$ \$2-6 $||$ \$4-1$\downarrow$ ) &4-2 &Concat(\$1-12$\uparrow$ $||$ \$2-6$\uparrow$ $||$ \$4-1 )\\
        \cline{2-7}
        &1-14 &Conv(192,64), ReLU &2-8 &Conv(192,64), ReLU &4-3 &Conv(192,64), ReLU\\
        \cline{2-7}
        &1-15 &RB$\times4$ &2-9 &RB$\times4$ &4-4 &RB$\times4$\\
        \cline{2-7}
        &1-16 &Conv(64,64) &2-10 &Conv(64,64) &4-5 &Conv(64,64)\\
        \cline{2-7}
        &1-17 &\$1-12 + \$1-16 &2-11 &\$2-6 + \$2-10 &4-6 &\$4-1 + \$4-5\\
\hline
\multirow{4}*{Stage4}  & & & & &4-7 &Concat(\$1-17$\uparrow$ $||$ \$2-11$\uparrow$ $||$ \$4-6)\\
\cline{6-7}
        & & & & &4-8 &Conv(192,64), ReLU \\
\cline{6-7}
        & & & & &4-9 &Conv(64,32) \\
\cline{6-7}
        & & & & &4-10 &Conv(32,3)\\
\hline
\end{tabular}
}
\end{center}
\end{table*}

\begin{table}[htb]
\caption{Network structure of the discriminator. Conv($N_{in}$, $N_{out}$, $S$) indicates the convolutional layer with $N_{in}$ input channels, $N_{out}$ output channels and stride $S$. The kernel size is $3\times3$ for all convolutional layers. The parameter is 0.2 for all the leaky ReLU layers. The size of HR and SR input is $160 \times 160 \times 3$}
\label{tab:discriminator}
\begin{center}
\begin{tabular}{|c|c|c|}
\hline
Id  &Layer Name \\
\hline
0 &Conv(3,32,1), LReLU \\
\hline
1 &Conv(32,32,2), LReLU \\
\hline
2 &Conv(32,64,1), LReLU \\
\hline
3 &Conv(64,64,2), LReLU \\
\hline
4 &Conv(64,128,1), LReLU \\
\hline
5 &Conv(128,128,2), LReLU \\
\hline
6 &Conv(128,256,1), LReLU \\
\hline
7 &Conv(256,256,2), LReLU \\
\hline
8 &Conv(256,512,1), LReLU \\
\hline
9 &Conv(512,512,2), LReLU \\
\hline
10 &FC(12800,1024), LReLU \\
\hline
11 &FC(1024,1)\\
\hline
\end{tabular}
\end{center}
\end{table}

\section{Texture Transformers on Different Scales}\label{sec:num_TT}
\setcounter{table}{0}
Our proposed TTSR contains three stacked texture transformers. The texture transformer at each scale fuses HR texture features of different levels from the Ref image. Here we conduct experiments of using the texture transformers on different scales. The model here is without CSFI since CSFI is designed for multi-scale stacked texture transformers. Table~\ref{tab:tt} shows the results. The larger scale the texture transformer is applied at, the more performance it brings, which demonstrates that the texture features at a larger scale have a less loss of details. When we gradually add the texture transformers at other scales, the performance can be further improved.

\begin{table}[htb]
\caption{Performance on CUFED5 testing set using texture transformers on different scales.}
\label{tab:tt}
\begin{center}
\begin{tabular}{|c|c|c|c|}
\hline
  scale$1\times$ &scale$2\times$ &scale$4\times$ &PSNR/SSIM\\
  \hline
  \checkmark & & &26.56 / .785 \\
  &\checkmark & &26.77 / .793 \\
  & & \checkmark &26.85 / .796 \\
  \checkmark &\checkmark & &26.80 / .793 \\
  \checkmark &\checkmark &\checkmark &26.92 / .797 \\
\hline
\end{tabular}
\end{center}
\end{table}

\section{Running Time and Model Size}\label{sec:time}
\setcounter{table}{0}
In this section, the running time and the model size of TTSR will be discussed. We compare the proposed TTSR with state-of-the-art SISR and RefSR approaches, RCAN~\cite{zhang2018image}, RSRGAN~\cite{zhang2019ranksrgan}, CrossNet~\cite{zheng2018crossnet} and SRNTT~\cite{zhang2019image}. For running time, all approaches are run on a Tesla V100 PCIe GPU and tested on an $83 \times 125 \times 3$ LR input image with the up-sampling factor of $4\times$. Table~\ref{tab:time_size} shows the comparison results. Specifically, the stacked texture transformers cost 0.037s and the other parts cost 0.059s, and TTSR takes a total time of 0.096s. The results show that TTSR achieves the best performance with a relatively small parameter number and running time.

\begin{table}[htb]
\caption{Running time and parameter number of different approaches. The last column shows the PSNR/SSIM performance on CUFED5 testing set.}
\label{tab:time_size}
\begin{center}
\begin{tabular}{|c|c|c|c|}
\hline
Approach &Time &Param. &PSNR/SSIM \\
\hline
RCAN~\cite{zhang2018image} &0.108s &16M &26.06 / .769\\
\hline
RSRGAN~\cite{zhang2019ranksrgan} &0.007s &1.5M &22.31 / .635\\
\hline
CrossNet~\cite{zheng2018crossnet} &0.229s &33.6M &25.48 / .764\\
\hline
SRNTT~\cite{zhang2019image} &4.977s &4.2M &26.24 / .784\\
\hline
TTSR &0.096s &6.4M &27.09 / .804\\
\hline
\end{tabular}
\end{center}
\end{table}

\section{More Visual Comparison}\label{sec:visual}
In this section, we show more comparison results among the proposed TTSR and other SR methods, including RDN~\cite{zhang2018residual}, RCAN~\cite{zhang2018image}, RSRGAN~\cite{zhang2019ranksrgan}, CrossNet~\cite{zheng2018crossnet} and SRNTT~\cite{zhang2019image}. RCAN has achieved state-of-the-art  performance on both PSNR and SSIM in recent years and RSRGAN is considered to achieve the state-of-the-art visual quality. CrossNet and SRNTT are two state-of-the-art RefSR approaches which significantly outperform previous RefSR methods. The visual comparison results on CUFED5~\cite{zhang2019image}, Sun80~\cite{sun2012super}, Urban100~\cite{huang2015single} and Manga109~\cite{matsui2017sketch} are shown in Figure~\ref{fig:cufed_0}-\ref{fig:cufed_3}, Figure~\ref{fig:sun80_0}-\ref{fig:sun80_1}, Figure~\ref{fig:urban100_0}-\ref{fig:urban100_1} and Figure~\ref{fig:manga109_0}-\ref{fig:manga109_1}, respectively.

\setcounter{figure}{0}
\begin{figure*}[t]
    \centering
    \includegraphics[width=1\linewidth, page=1]{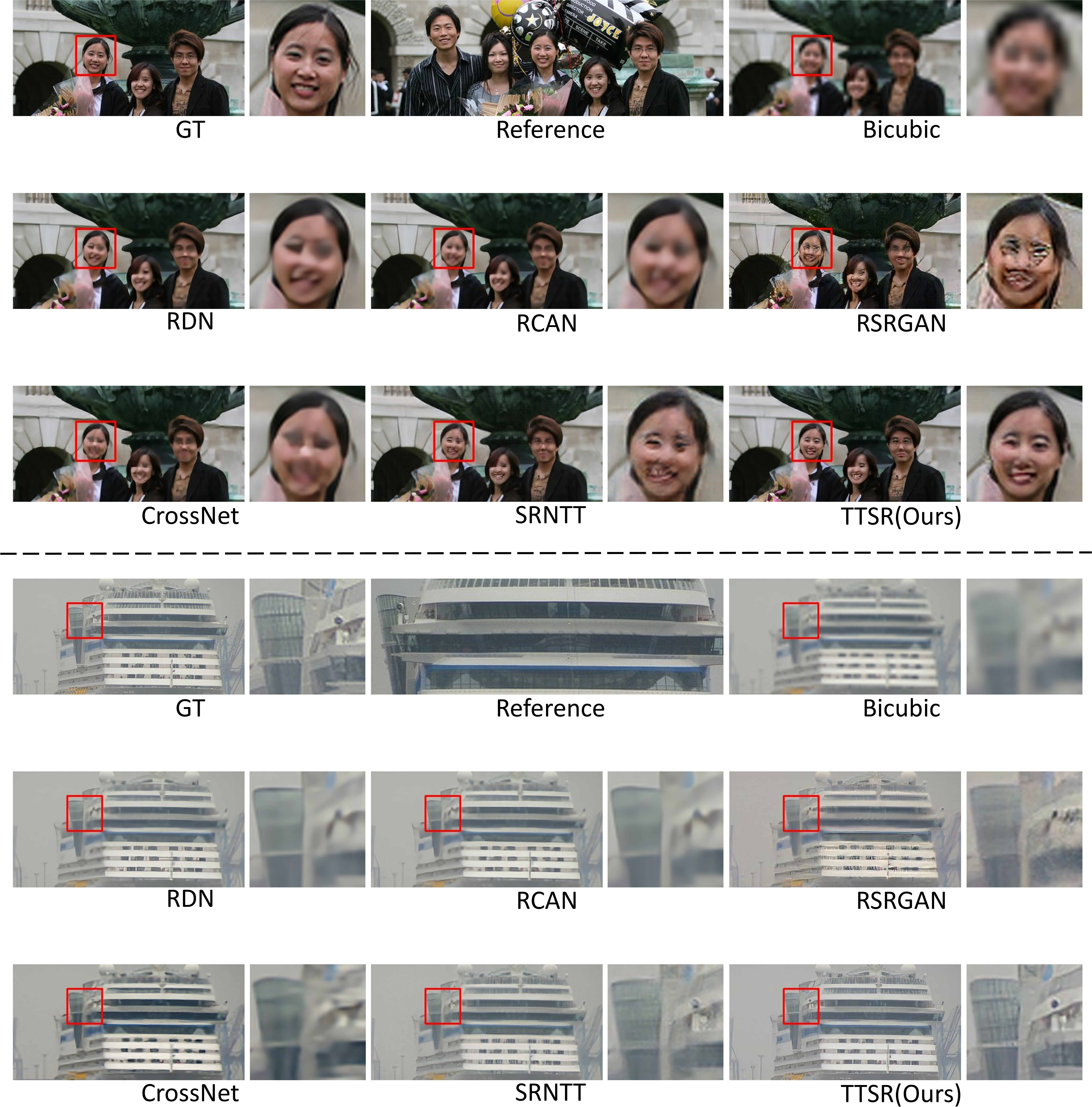}
    \caption{Visual comparison of different SR methods on CUFED5~\cite{zhang2019image} dataset.}
    \label{fig:cufed_0}
\end{figure*}

\begin{figure*}[t]
    \centering
    \includegraphics[width=1\linewidth, page=2]{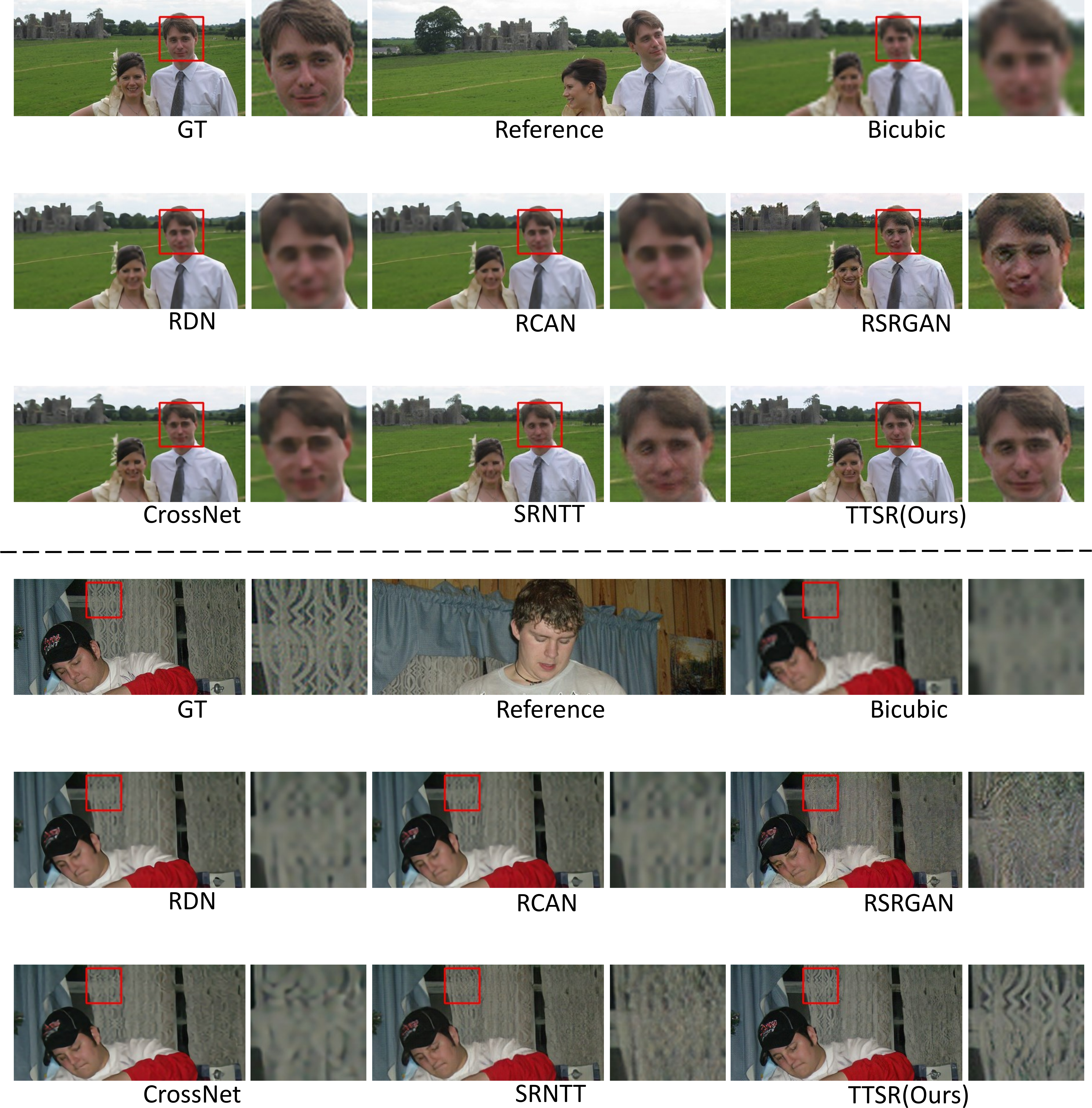}
    \caption{Visual comparison of different SR methods on CUFED5~\cite{zhang2019image} dataset.}
    \label{fig:cufed_1}
\end{figure*}

\begin{figure*}[t]
    \centering
    \includegraphics[width=1\linewidth, page=3]{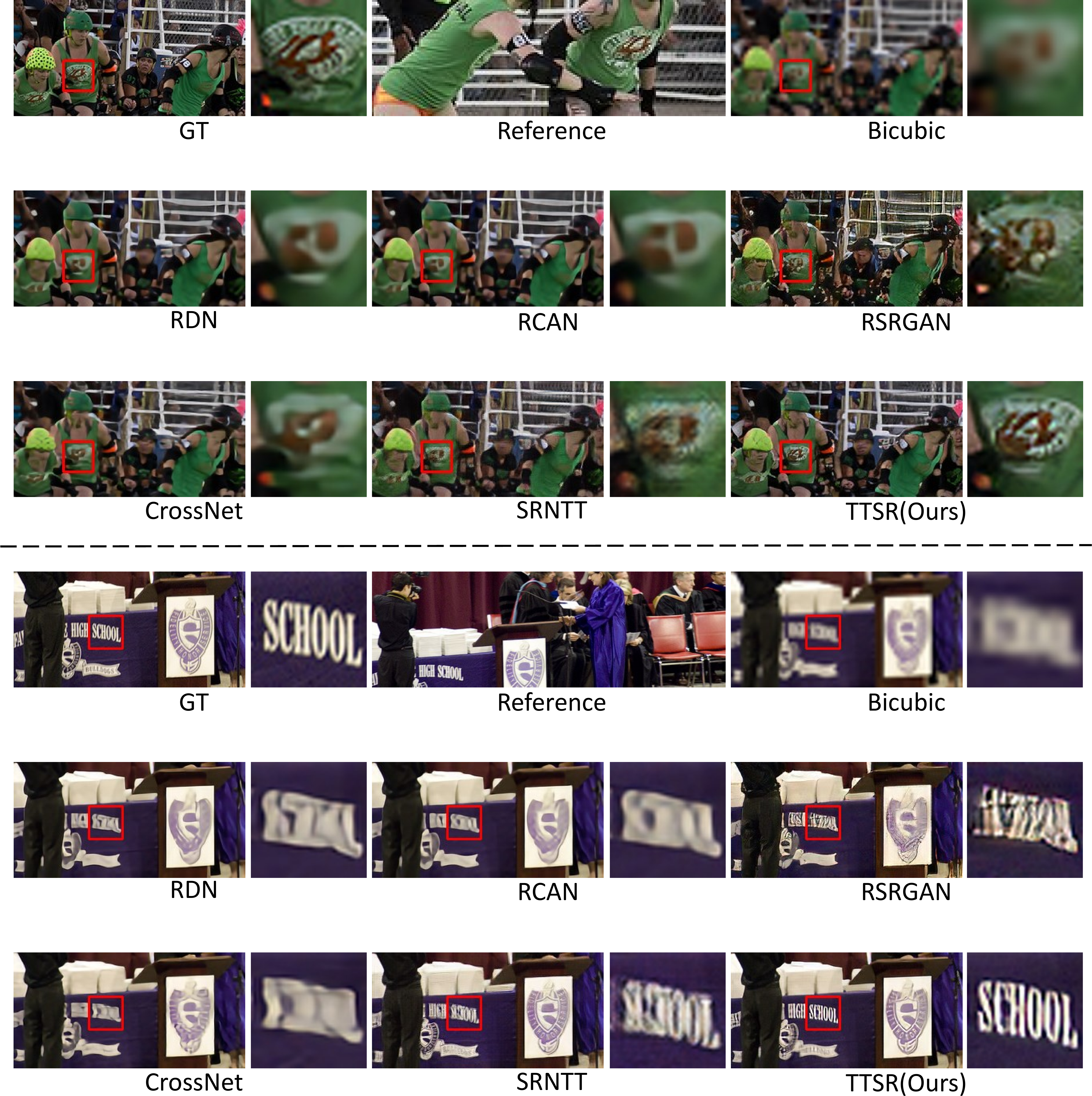}
    \caption{Visual comparison of different SR methods on CUFED5~\cite{zhang2019image} dataset.}
    \label{fig:cufed_2}
\end{figure*}

\begin{figure*}[t]
    \centering
    \includegraphics[width=1\linewidth, page=4]{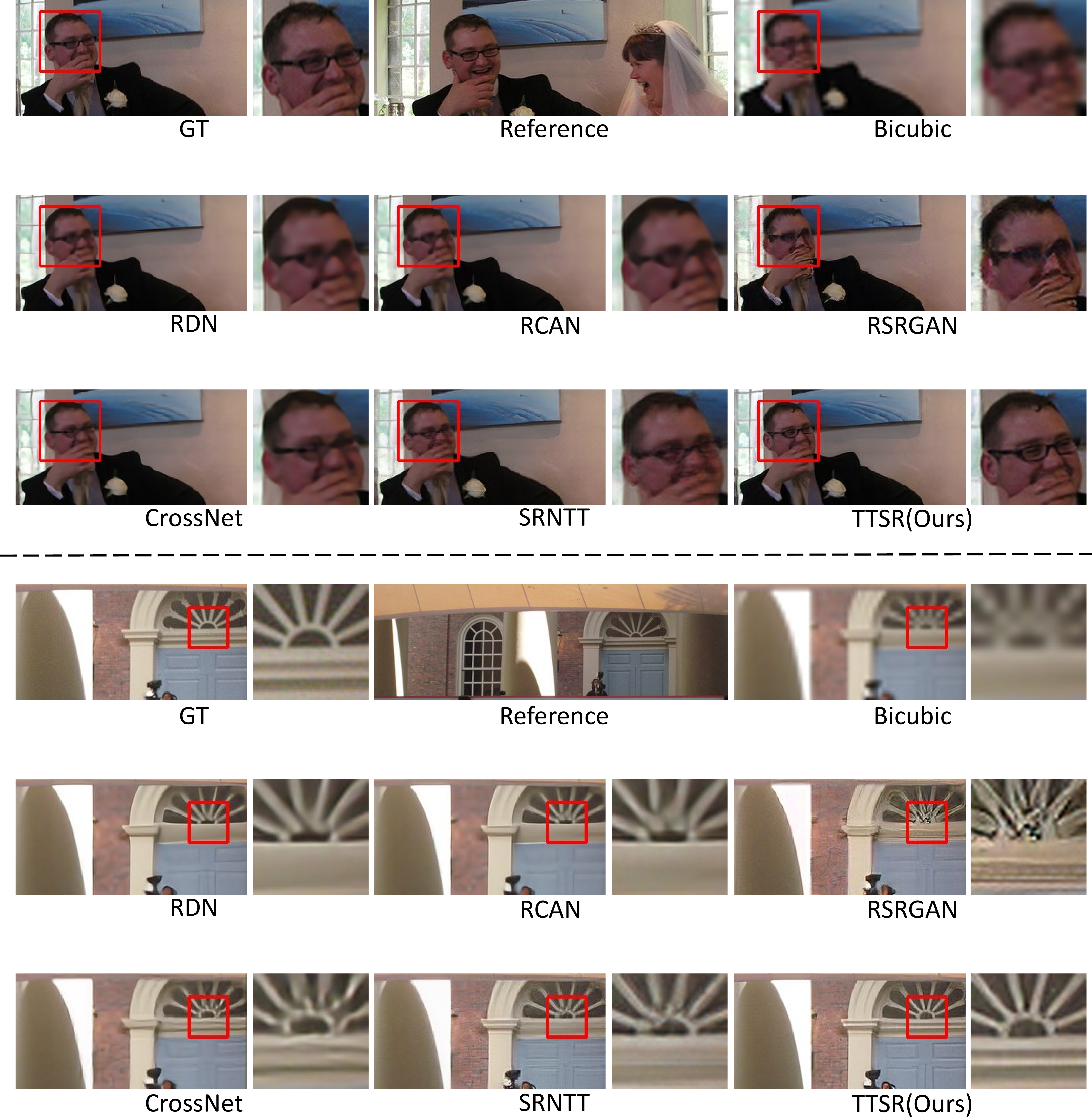}
    \caption{Visual comparison of different SR methods on CUFED5~\cite{zhang2019image} dataset.}
    \label{fig:cufed_3}
\end{figure*}

\begin{figure*}[t]
    \centering
    \includegraphics[width=1\linewidth, page=5]{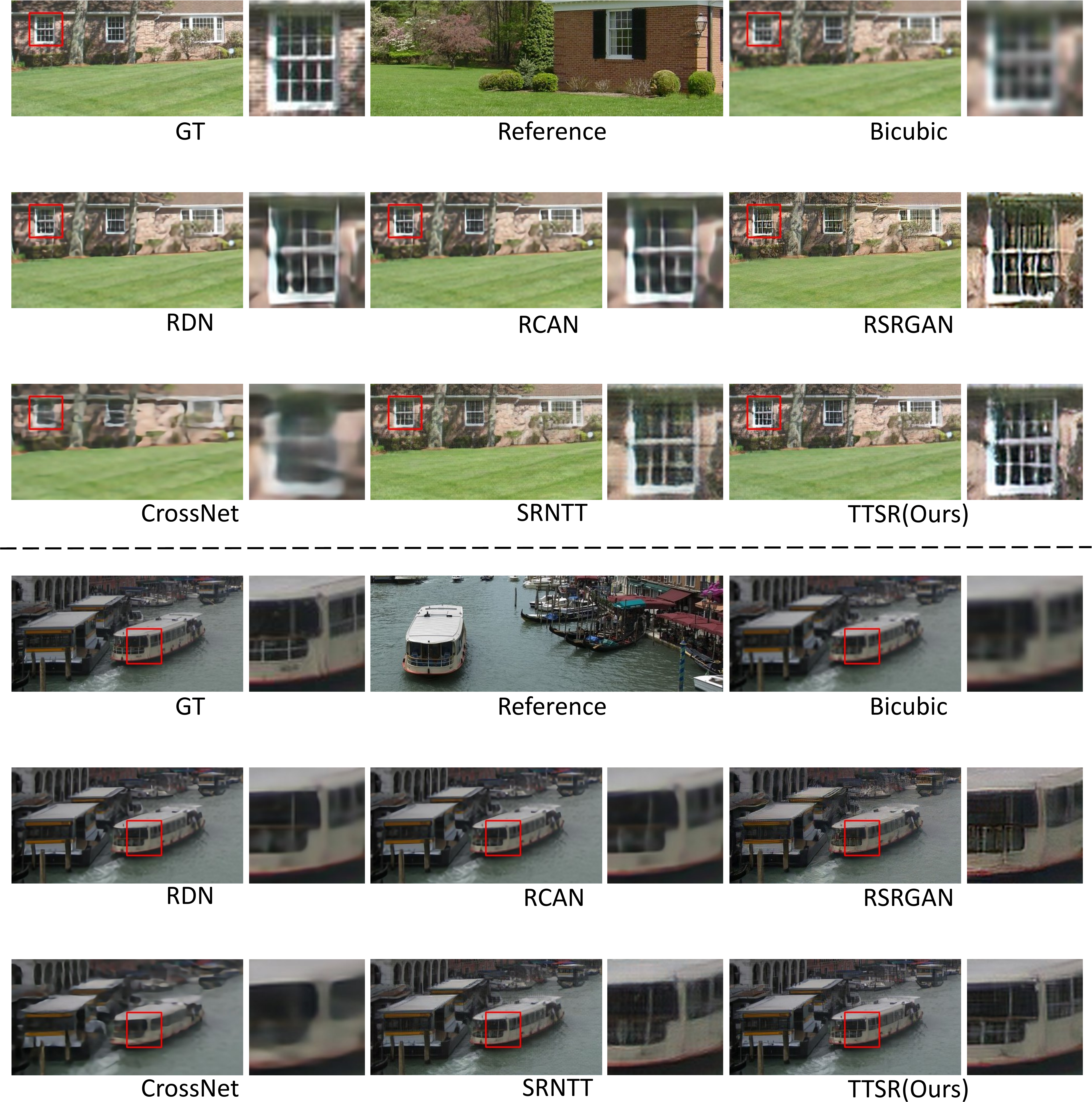}
    \caption{Visual comparison of different SR methods on Sun80~\cite{sun2012super} dataset.}
    \label{fig:sun80_0}
\end{figure*}

\begin{figure*}[t]
    \centering
    \includegraphics[width=1\linewidth, page=6]{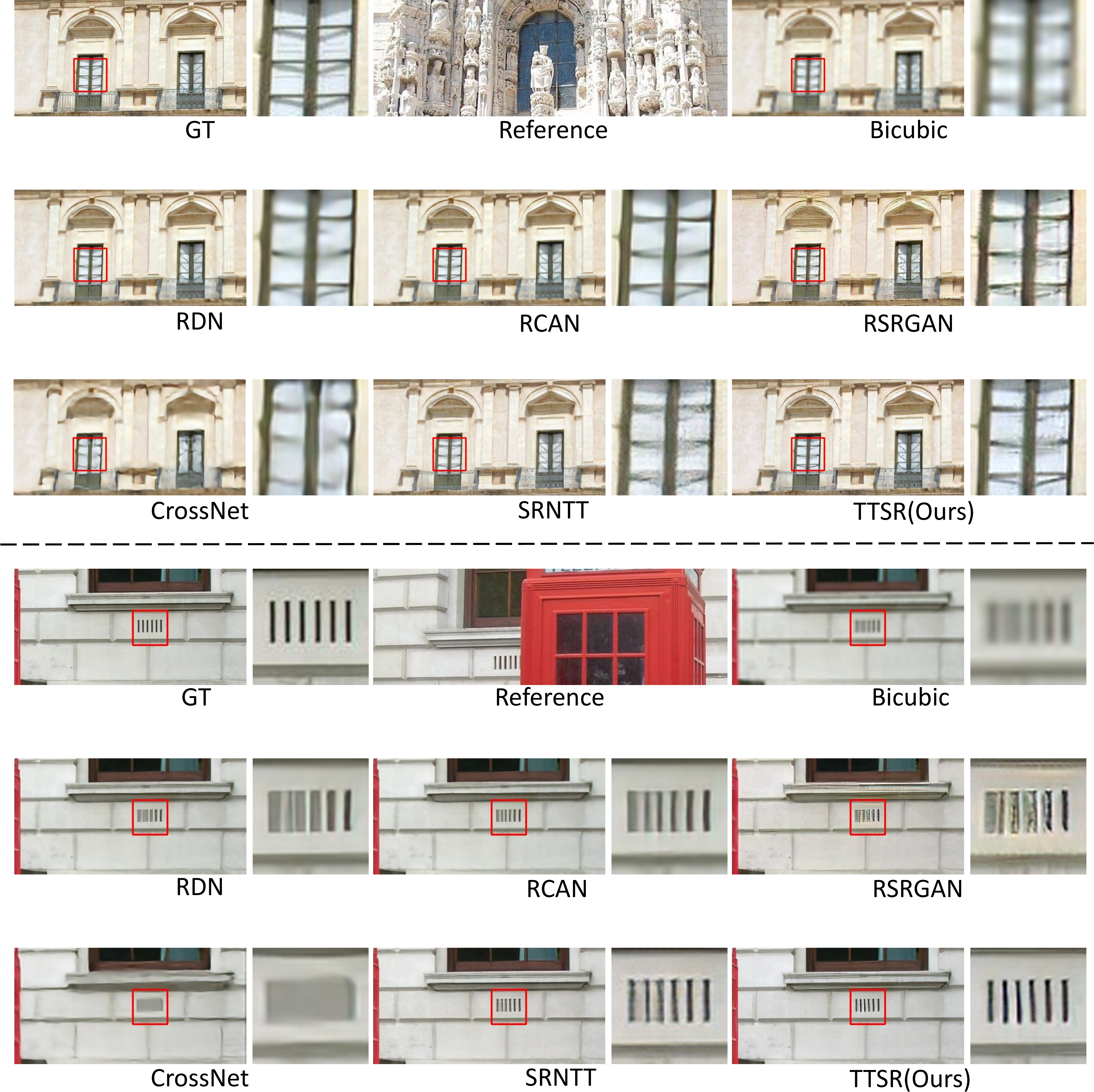}
    \caption{Visual comparison of different SR methods on Sun80~\cite{sun2012super} dataset.}
    \label{fig:sun80_1}
\end{figure*}

\begin{figure*}[t]
    \centering
    \includegraphics[width=1\linewidth, page=7]{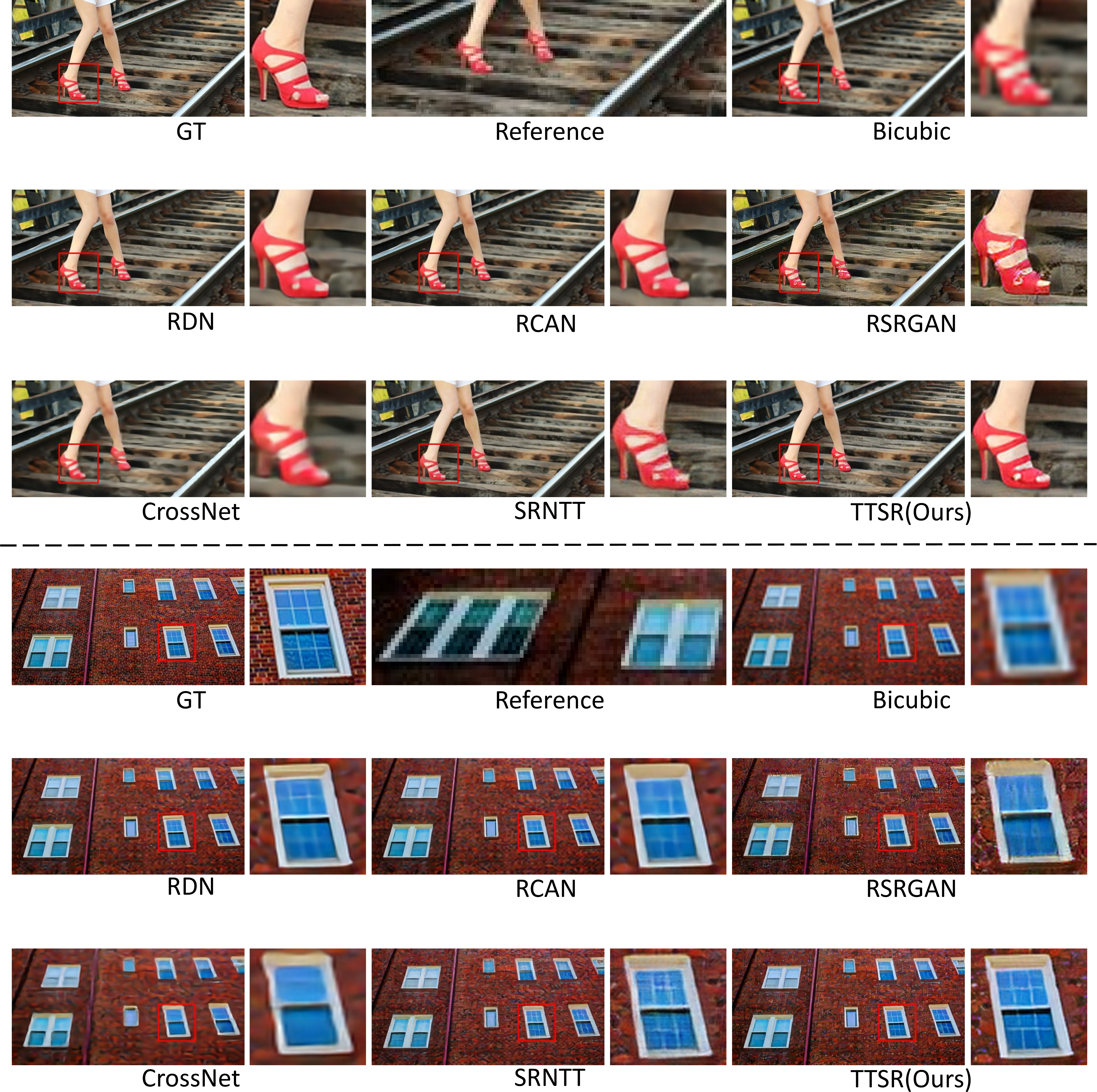}
    \caption{Visual comparison of different SR methods on Urban100~\cite{huang2015single} dataset.}
    \label{fig:urban100_0}
\end{figure*}

\begin{figure*}[t]
    \centering
    \includegraphics[width=1\linewidth, page=8]{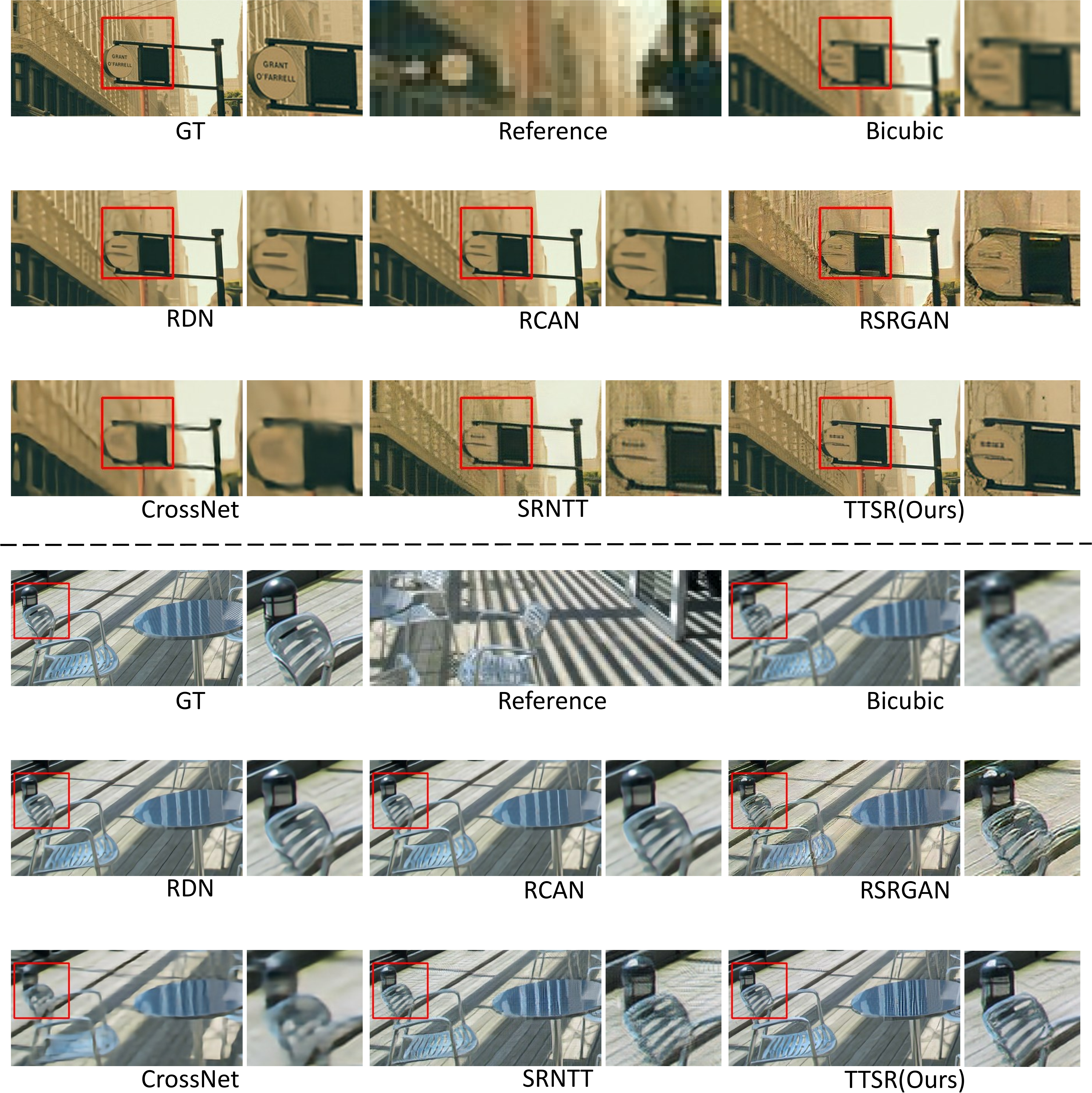}
    \caption{Visual comparison of different SR methods on Urban100~\cite{huang2015single} dataset.}
    \label{fig:urban100_1}
\end{figure*}

\begin{figure*}[t]
    \centering
    \includegraphics[width=1\linewidth, page=9]{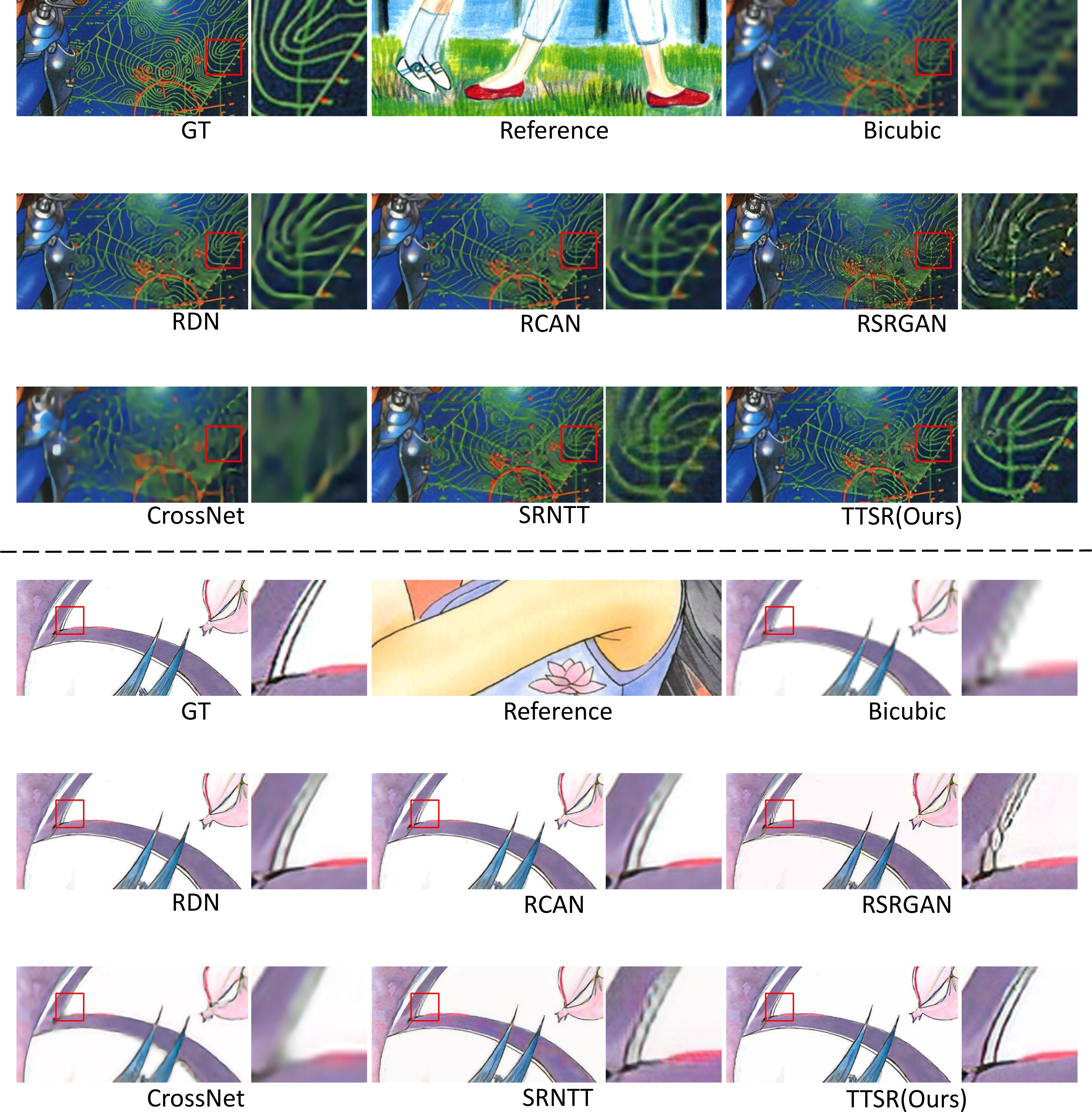}
    \caption{Visual comparison of different SR methods on Manga109~\cite{matsui2017sketch} dataset.}
    \label{fig:manga109_0}
\end{figure*}

\begin{figure*}[t]
    \centering
    \includegraphics[width=1\linewidth, page=10]{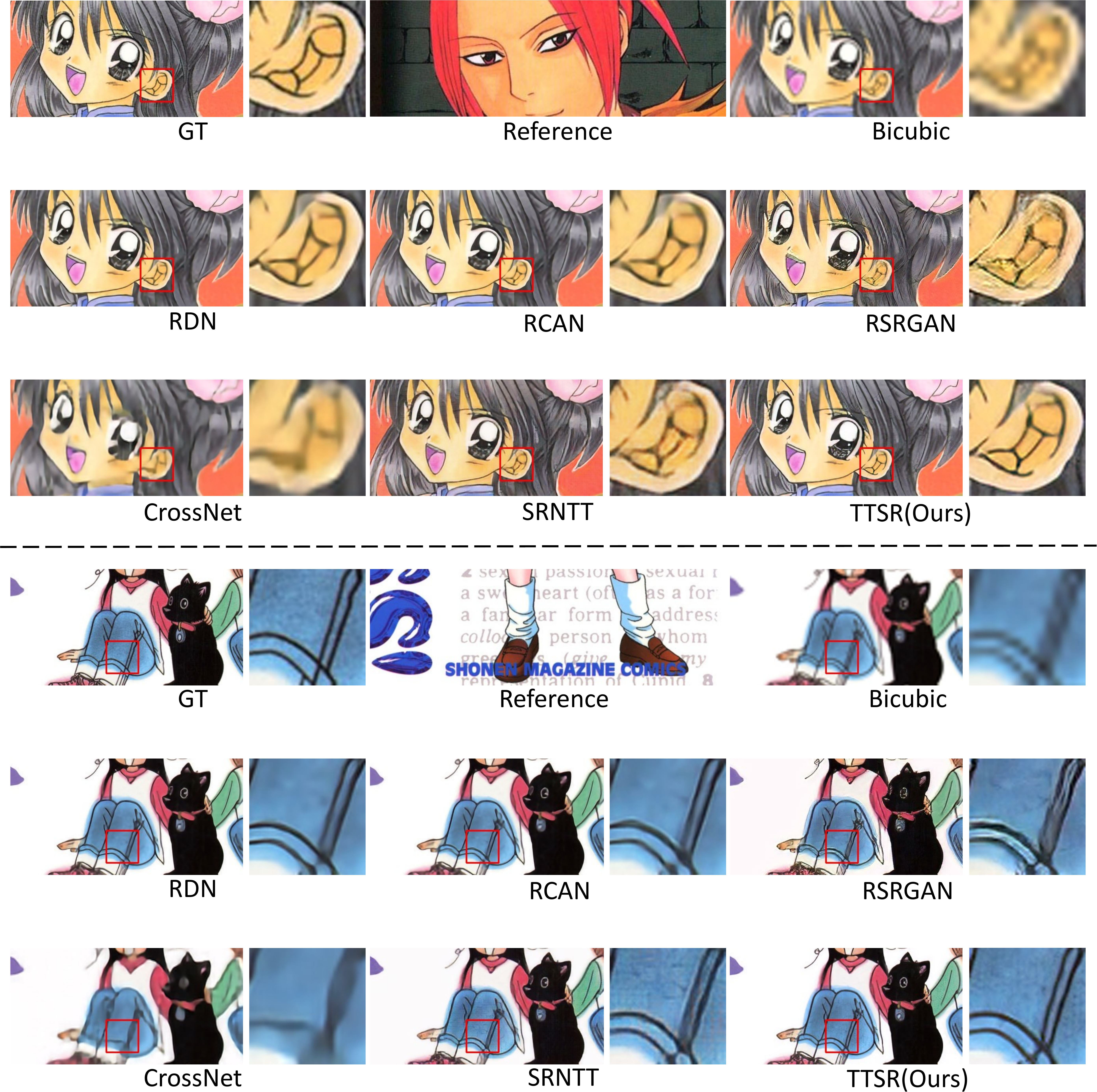}
    \caption{Visual comparison of different SR methods on Manga109~\cite{matsui2017sketch} dataset.}
    \label{fig:manga109_1}
\end{figure*}

\end{document}